\documentclass{article}

\usepackage[preprint]{neurips_2020}

\usepackage[utf8]{inputenc} 
\usepackage[T1]{fontenc}    
\usepackage{hyperref}       
\usepackage{url}            
\usepackage{booktabs}       
\usepackage{amsfonts}       
\usepackage{nicefrac}       
\usepackage{microtype}      

\usepackage{amsmath}

\usepackage{graphicx} 
\graphicspath{{figures/}}
\usepackage{natbib}

\usepackage{chngcntr}

\counterwithout{figure}{section}

\renewcommand{\thefigure}{\arabic{section}.\arabic{figure}}

\renewcommand{\thetable}{\arabic{section}.\arabic{table}}

\title{Self-supervised learning of multi-omics embeddings \\ in the low-label, high-data regime}

\author{%
  Christian John Hurry\thanks{Corresponding author (https://gsk.ai/fellowships/)}\\
  GSK.ai\\
  \texttt{christian.j.hurry@gsk.com} \\ \And Emma Slade \\
  GSK.ai\\
  \texttt{emma.x.slade@gsk.com} \\
}

\begin{document}

\maketitle

\begin{abstract}

Contrastive, self-supervised learning (SSL) is used to train a model that predicts cancer type from miRNA, mRNA or RPPA expression data. This model, a pretrained FT-Transformer, is shown to outperform XGBoost and CatBoost, standard benchmarks for tabular data, when labelled samples are scarce but the number of unlabelled samples is high. This is despite the fact that the datasets we use have $\mathcal{O}(10^{1})$ classes and $\mathcal{O}(10^{2})-\mathcal{O}(10^{4})$ features. After demonstrating the efficacy of our chosen method of self-supervised pretraining, we investigate SSL for multi-modal models. A late-fusion model is proposed, where each omics is passed through its own sub-network, the outputs of which are averaged and passed to the pretraining or downstream objective function. Multi-modal pretraining is shown to improve predictions from a single omics, and we argue that this is useful for datasets with many unlabelled multi-modal samples, but few labelled unimodal samples. Additionally, we show that pretraining each omics-specific module individually is highly effective. This enables the application of the proposed model in a variety of contexts where a large amount of unlabelled data is available from each omics, but only a few labelled samples.

\end{abstract}

\section{Introduction}

Self-supervised learning (SSL) leverages large unlabelled datasets to learn meaningful representations of data and has been used to achieve strong performance in supervised problems with limited labelled data \citep{balestriero2023cookbook}. In SSL a model is trained on a pretext task, the targets of which are typically generated by augmenting the training samples. In computer vision and natural language processing, examples of pretext tasks include predicting which of a finite set of rotations has been applied to an image \citep{gidaris2018unsupervised} or predicting words that have been masked from a sequence of text \citep{devlin2018bert}. Self-supervised methods can also be based on reconstructive objectives, such as masked autoencoders, which take samples with features randomly masked out as input and are trained to reconstruct the values of those masked features \citep{he2022masked,geng2022multimodal,pathak2016context}. Contrastive learning is another popular self-supervised method which assumes knowledge of which samples should have similar or different representations, such that a model can be trained to align/contrast the representations of different samples \citep{chen2020simple,chen2020improved,he2020momentum,oord2018representation}. Training a model on sufficiently difficult pretraining tasks allows a model to learn useful representations from unlabelled data. After pretraining, a model can be attached to a new module which takes the representations as input and is trained on the task of interest, referred to as the downstream task. The pretrained part of the model, referred to as the backbone, can either be frozen or finetuned along with the new layers for the downstream task. SSL has shown to be particularly useful in the \textbf{low-label regime} where the number of unlabelled samples dwarfs the number of labelled samples \citep{bengar2021reducing}.

While SSL has been successful in computer vision and natural language processing, it has only recently been applied to the tabular domain, where deep learning remains contentious due to the strong performance of gradient boosted decision trees (GBDTs)\citep{gorishniy2021revisiting,grinsztajn2022tree}. Self-supervised methods rely on augmentation to generate targets and to form sensible pretext tasks, the choice of which can affect performance on downstream tasks. Images and language have structure that provides intuitive augmentation and pretext tasks. For example, a model that is trained to predict a rotation applied to an image will have to learn features of the classes of the images in the training set to distinguish between upright and downturned orientations. Tabular data, on the other hand, often lacks this explicit structure and hence it remains an open question how to augment tabular data and define pretext tasks which will learn useful representations for downstream tasks. Approaches so far have considered both reconstructive \citep{yoon2020vime,arik2021tabnet,ucar2021subtab} and/or contrastive objectives \citep{bahri2021scarf,somepalli2021saint,onishi2023rethinking} and have shown that self-supervised pretraining for tabular data can improve model performance in the low-label regime. 

In computer vision and NLP, it has become common practice to pretrain very large models on very large datasets, such as ImageNet \citep{deng2009imagenet} and Common Crawl, and then to finetune these models on smaller, domain-specific, datasets for the task of interest. By pretraining on a different dataset, that is much larger than the domain-specific dataset of interest, the model can learn more effectively from the small dataset of interest, similar to how SSL is used to achieve good performance with few labels. For the tabular domain, this is not standard practice. Images and language have structural information that can be learned from generic image and language datasets. Tabular data typically lacks such structure. An arbitrary pair of tabular datasets will usually have few features in common. Hence, pretraining on a large tabular dataset and finetuning on a smaller domain-specific dataset, is not usually possible. 

Expression data from the transcriptome and proteome measured by various omics-based technologies, however, is often stored in tabular format and is more structured than typical tabular data. For example, the data collected from two experiments using the same RNAseq machine and protocol will typically contain a high overlap in the features that are measured. Hence, it may be that large tabular datasets of expression data can be collated, and used for pretraining large models that may be useful in contexts where there is limited labelled data for the domain-specific task of interest. Limitations on the amount of labelled data available are typical of the biomedical domain, due to the cost of acquiring labels through experimentation or clinical trials. Therefore, developing models that can leverage large amounts of unlabelled data to learn effectively from few labels remains an important problem. 

An additional problem to consider is the integration of data collected from multiple omics-based technologies. The advent of high throughput screening has led to the collection of large multi-omics datasets, wherein samples are comprised of features from multiple omics-based modalities. One would expect that in developing models that make predictions related to the understanding of complex diseases, that it would be beneficial to utilise information from a combination of molecular sources. To this end, deep neural networks have been developed to integrate data from multiple omics-technologies and have shown that representations learned from a combination of omics data have stronger downstream performance on tasks such as cancer sub-type prediction and survival prediction, than representations learned from the same data using just a single omics source \citep{zhang2021omiembed}. 

The application of self-supervised methods to expression data has only recently been considered. Works which have applied SSL to expression data typically concern the integration of multiple omics sources and learning strong representations with limited supervision. Notably, the majority of this work has considered contrastive objectives. Contrastive objectives have been shown to optimise for alignment of similar samples such that they are close on the unit hypersphere, while also optimising for maximal information, by distributing samples uniformly across the unit hypersphere \citep{wang2020understanding}. When applied to high-dimensional data, contrastive learning does not rely on a large output space, unlike reconstruction methods that are, hence, prone to overfitting for medium scale datasets. Contrastive learning has recently been applied to single cell expression datasets \citep{han2022self,yang2022contrastive,liu2023single}. Collectively, these works show that contrastive learning provides strong representations with no or little supervision, and that a contrastive objective effectively integrates information from multiple omics sources, whilst also making the model robust to batch effects. A combination of contrastive and reconstructive self-supervised objectives have been applied to bulk multi-omics datasets, where measurements are from a population of cells as oppose to single cells \citep{hashim2022self,zhao2023clclsa}.

An additional problem with developing models that integrate data from multiple omics, is that multi-omics datasets commonly contain samples which are missing an entire modality, or multiple modalities. Recent work has suggested methods to handle data with an arbitrary combination of a set number of modalities, by either generating missing modalities with an encoder-decoder module \citep{zhao2023clclsa}, or by learning a shared latent space common to all modalities \citep{lee2021variational}.

In our work, we apply a recent self-supervised method for tabular data, that uses an attention-based architecture, to multi-omics expression data \citep{onishi2023rethinking}. This method uses a contrastive objective, which aligns the latent representation of a sample and an augmented sample. This work is notable for the augmentation it suggests for tabular data: a random set of features are masked, and masked features are imputed with a common value, which is a learned parameter of the model. This augmentation, and its use with contrastive pretraining, is referred to as \textbf{Mask Token Replacement} (MTR). In this work, we show that MTR is an effective self-supervised pretraining scheme for expression data. We show that MTR learns useful representations that perform well in the low-label regime when compared with GBDTs, which are strong baselines when working with few samples of high-dimensional data. We also find that pretraining with MTR allows a model to perform well even when an extreme number (75\%) of features are missing at test time. Missing features are a common batch effect associated with sequencing technologies, and contrastive learning has previously been shown to mitigate for this in single cell studies \citep{han2022self}.

We then propose a novel architecture for multi-omics integration, which proposes to combine omics-specific transformers whose latent representations are averaged for later downstream use. We demonstrate the MTR learns effective multi-omics representations, and that MTR successfully integrates information from a combination of multiple omics sources, outperforming other contrastive objectives based on aligning the representation of different modalities \citep{hager2023best,radford2021learning}. We argue the case for late-fusion multi-modal models, by demonstrating that the modularity of our proposed architecture allows for greater flexibility in how it is trained, and to what datasets it can later be tested against. For instance, we show that we can extract an omics-specific module from our pretrained multi-modal model, and find that it produces stronger predictions from a single omics than an identical model that has been pretrained and finetuned on the same data, but with features from a single omics only. We also show that our model can be pretrained in a joint-manner, which requires a large number of multi-modal samples, or by pretraining each omics-specific module individually. The latter means that our multi-modal model can be pretrained with large amounts of unlabelled data when there are few samples where both modalities have been measured. These features of our approach enable us to handle samples with missing modalities during training or test time without relying on generative/reconstructive methods, which for high-dimensional data require large output spaces and can be prone to overfitting for medium scale datasets.

\section{TCGA Pan-Cancer Atlas}

In this work, we developed self-supervised methods for expression data, using data from the cancer genome atlas (TCGA)\footnote{Publicly available data (https://www.cancer.gov/ccg/research/genome-sequencing/tcga/using-tcga-data/citing).}. The Pan-Cancer dataset from TCGA contains data with measurements from the genome, transcriptome, epigenome, and proteome. In total, it contains samples from 11,000 tumours, each from 1 of 33 sub-classes of cancer. From the original TCGA Pan-Cancer Atlas dataset, we created 6 subsets, each of which we considered as a separate dataset. These datasets are summarised in Table \ref{tab: data summary}, and we detail how they were accessed in Appendix \ref{app: data}. Each dataset contains data from a single omics source, or a pair of omics sources. We considered miRNA expression, mRNA expression, reverse phase protein array (RPPA) data, and paired combinations of these data types. For each dataset, we removed less represented classes, only keeping classes with $>100$ samples. This allowed us to investigate the performance of our models in the low-label regime while keeping all data splits stratified, to preserve any class imbalance in the datasets.  
In addition to removing less represented classes, we also removed any samples with missing features - although we later investigated the effect of missing features, by introducing missingness synthetically. The datasets containing data from a pair of omics-sources only contained samples which have data from both modalities. Similarly, we later investigated the impact of missing-modalities later by introducing missing-modalities synthetically.  In all cases, the datasets were multi-class with significant class imbalance. A particular challenge of working with mRNA data is that the number of features is much greater than the number of samples. We evaluate our models by training them to predict which type of cancer a sample is from, and assess performance across each of our datasets.

\begin{table}[t]
\centering
  \caption{Summary of the datasets used to evaluate the models in this work. Each dataset is a subset of the Pan-Cancer Atlas. All classes in the datasets have >100 samples, and samples with missing features have been removed.}
  \label{tab: data summary}
\begin{tabular}{lccccl}\toprule 

           & No. Samples &  No. Features &   No. Classes  \\ \midrule
\textbf{mRNA}  & 8815 & 20531  & 22 \\
\textbf{miRNA} & 10330 & 743 & 23 \\
\textbf{RPPA }& 7204 & 210 & 23 \\ 
\midrule
\textbf{mRNA+miRNA}  & 8220  & 21276 & 20 \\
\textbf{mRNA+RPPA} & 5403 & 20743 & 17 \\
\textbf{miRNA+RPPA} & 6616 & 953 & 21 \\ 
 \bottomrule
\end{tabular}
\end{table}

\section{Masked tabular transformers}

In this work we train transformers, specialised for tabular data, via a contrastive learning objective implemented in recent work \citep{onishi2023rethinking}. As in this previous work, the backbone of our model is the FT-Transfromer (FTT), an encoder-only transformer which tokenises both categorical and numerical features (the latter via linear layers) \citep{gorishniy2021revisiting}. Numerical features in a tabular dataset are tokenised during a forward pass of the model, with each numerical feature being passed through its own linear layer, such that a continuous feature is transformed into a $d$-dimensional token (i.e vector) where $d$ is the number of units in the linear layer. During a forward pass, all features are tokenised, such that the input with $M$ total features is transformed to a stack of $M$ tokens of dimension $d$. An additional, randomly initialised, token - referred to as the class token - is appended to the stack of tokens. The full stack of tokens is then passed through a series of encoder-only transformer layers \citep{vaswani2017attention}. As the class token attends to all other tokens when passed through the transformer layers, it should retain information about all other tokens. Hence, linear layers may be attached to the class token after it has passed through the transformer layers, with a final output layer for the task associated with the training objective. 

\begin{figure}
  \centering
\includegraphics[width=0.65\linewidth]{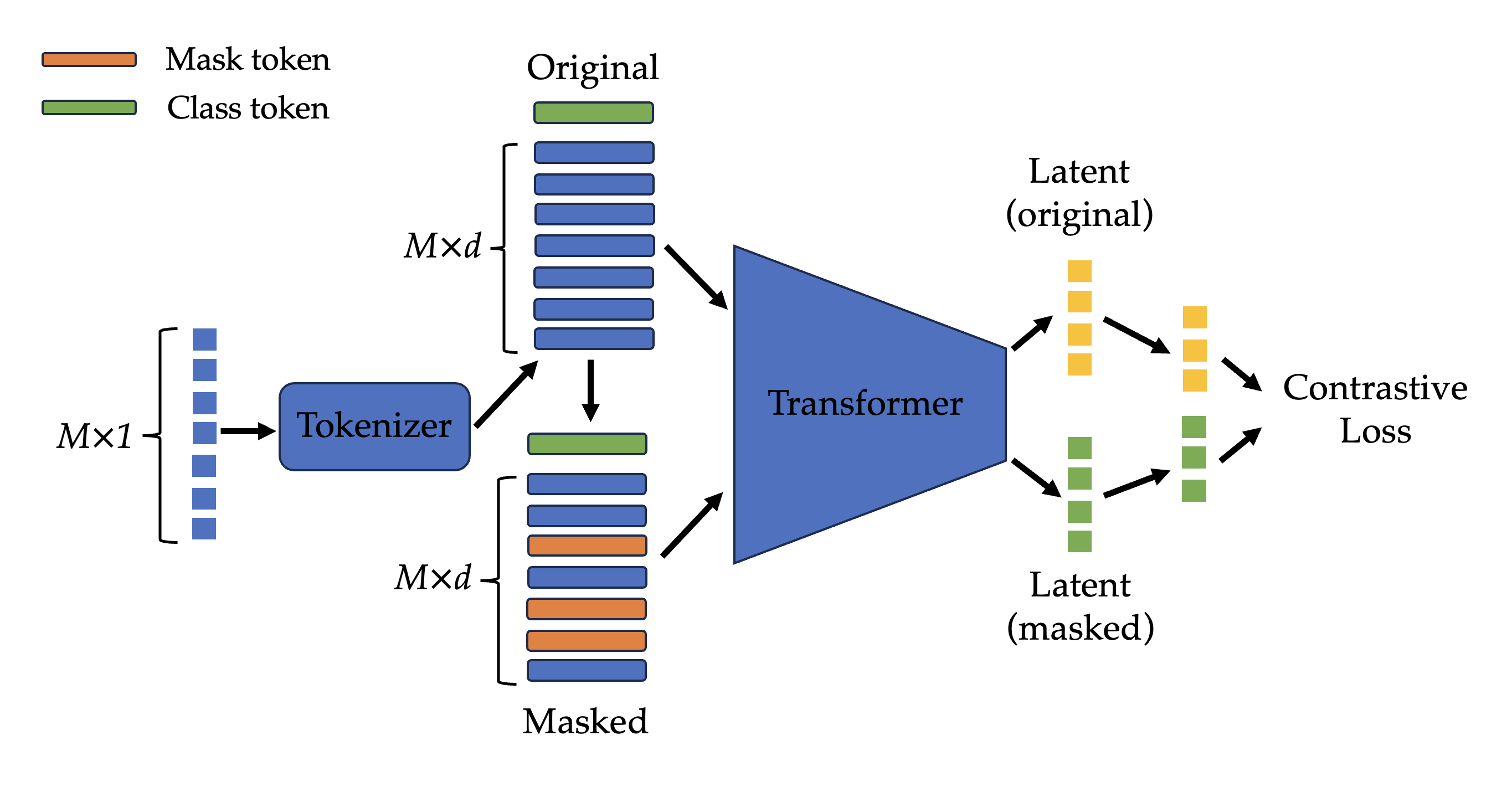}
  \caption{Sketch of contrastive learning with MTR. A sample with $M$ features is tokenised, such that each feature is transformed to a $d$-dimensional vector. The stack of $M$ tokens, with an additional class token, is passed through a series of transformer layers, which produces a latent representation of the input. From the original stack of tokens, a random fraction are replaced with the mask token. The new stack of tokens is then passed through the transformer to produce a second latent representation, corresponding to the masked input. Both latent representations are projected through linear layers. These projected representations are then aligned via a contrastive loss function. }
  \label{fig: mtr}
\end{figure}

We illustrate this architecture during contrastive pretraining in Figure \ref{fig: mtr}. During contrastive pretraining the class token, a latent representation of the input, is projected via linear layers. For each sample, the original set of tokens is passed through the model to yield one projected latent representation of the input data. The same sample is then passed through the model again, but after tokenisation, a random set of tokens are selected and replaced with a mask token. The mask token is a $d$-dimensional vector, and is a learned parameter of the model, initialised at random with values from the uniform distribution. A second projected latent representation is obtained as the new stack of tokens (including the mask tokens) is passed through the remaining layers of the model. If we consider $N$ to be the size of a mini-batch, $\mathbf{z}_i$ to be the projected latent representation of an original stack of tokens of sample $i$, and $ \mathbf{\tilde{z}}_{i}$ the projected latent representation of the masked stack of tokens, the model is trained to minimise the NTXent loss function given by,
\begin{equation}
\mathcal{L}_{contrastive} = - \sum_{i=1}^{N} \log \frac{\exp \left(sim\left( \mathbf{z}_{i},\mathbf{\tilde{z}}_{i} \right) / \tau \right)}{ \sum_{j=1}^{N}(1- \delta_{j,k}) \left[\exp \left( sim\left( \mathbf{z}_{j},\mathbf{z}_{k} \right) / \tau \right) + \exp \left( sim\left( \mathbf{z}_{j},\mathbf{\tilde{z}}_{k} \right) / \tau \right)\right]}.
\end{equation}
Here we have defined the Kronecker-Delta function $\delta_{i,j} =1 \forall i=j$ and 0 otherwise, and $sim(\mathbf{u},\mathbf{v}) = \mathbf{u}^{T} \mathbf{v} / || \mathbf{u} ||  ||\mathbf{v} || $ to be the cosine similarity between two vectors. Additionally, $\tau$ is a `temperature' parameter that is set to $\tau=1$ throughout this work.   
The model is trained to minimise the cosine distance between the latent representations corresponding to the original and masked set of tokens, whilst maximising the cosine distance between the latent representation (corresponding to both masked and unmasked sets of tokens) of the other samples in the batch. 
 
By training the model to minimise the cosine distance between the projected latent representations of the the original and masked set of tokens from the same sample, whilst maximising the distance with the latent representations of other samples, the model learns that samples with a fraction of tokens replaced with the mask token should produce similar representations to samples which are similar with respect to the unmasked features. Since the mask token is a learned parameter of the model, it learns the best imputation to align the representations, whilst distancing itself from other samples.
 
After pretraining, our model is then finetuned in a supervised manner. The linear layers attached to the class token during pretraining are removed and replaced by new linear layers, the last layer containing a number of outputs corresponding to the number of classes. The entire model is then finetuned to minimise the cross-entropy loss using the class labels.

\subsection{Masked transformers are effective in the low-label regime}

To assess the efficacy of mask token replacement (MTR) as a self-supervised pretraining method, we compare the performance of an FTT, with and without pretraining, on the cancer classification task. 20\% of the data was held out for the test set. The remaining data was then split (90:10) into a train and validation set. All of the training set was used for pretraining, but either 1\%, 5\% or 10\% of the training set was used for finetuning. All splits were stratified to keep the proportion of classes preserved in the splits. After splitting, standardisation and principle component analysis (PCA) was applied to the training set to reduce the dimensionality to 200 features.These transformations were then applied to the test and validation sets, to avoid indirect data leakage.  Standardisation and PCA was applied in this way in all other experiments described in this work, unless explicitly stated otherwise. The pretraining phase was set for 200 epochs. Both the pretrained and randomly initialised FTT were finetuned for a maximum of 200 epochs, using the loss on the validation set as an early stopping criteria. A patience parameter of 10 epochs was added to the early stopping criteria to prevent underfitting. 

We compared the overall accuracy, the macro-averaged AUROC, and the macro-averaged F1 score and precision, of the models on the test set in Table \ref{tab: FT vs MTR}. We find that there is a significant improvement in the performance of the model when pretrained using MTR. The improvement however reduces as the size of the finetuning dataset is increased. We surmise that for datasets where the number of unlabelled samples is much larger than the number of labelled samples, self-supervised pretraining via MTR can yield significant improvements in performance.

\begin{table}[t]
  \caption{Performance on the cancer classification task of an FTT, with or without pretraining, trained on the \textbf{miRNA} dataset. Results shown for different proportions of the training set used for finetuning. Test metrics shown are overall accuracy, and the macro-averaged AUROC, F1 score, and precision. Test metrics are averaged over 5 seeds with (±) indicating the standard deviation. }
  \label{tab: FT vs MTR}
  \small
\begin{tabular}{lccccccl}\toprule
    \% training set &       & Accuracy & AUROC & F1   & Precision  \\ \midrule
 \textbf{1\%}&FTT  & 0.6340 ± 0.0180 &0.8933± 0.0114& 0.5078 ± 0.0313 & 0.5462 ± 0.0369\\ 
 &MTR & 0.7466 ± 0.0227 & 0.9158 ± 0.0121& 0.6345 ± 0.0248 &  0.6564 ± 0.0323\\ \midrule
  \textbf{5\%}&FTT  & 0.8471 ± 0.0098 & 0.9452 ± 0.0092 & 0.7667 ± 0.0156 & 0.7796 ± 0.0131  \\
 &MTR &0.8772 ± 0.0107  & 0.9502 ± 0.0104 & 0.7989 ± 0.0124 & 0.8102 ± 0.0145 \\ \midrule
  \textbf{10\%}&FTT  & 0.8867 ± 0.0129 & 0.9528 ± 0.0085 &0.8100 ± 0.0153  & 0.8214 ± 0.0135 \\
 &MTR & 0.8992 ± 0.0115 & 0.9538 ± 0.0081 & 0.8236 ± 0.0161 &  0.8340 ± 0.0162 \\
 \bottomrule
\end{tabular}
\end{table}

\begin{figure}
  \centering
\includegraphics[width=0.45\linewidth]{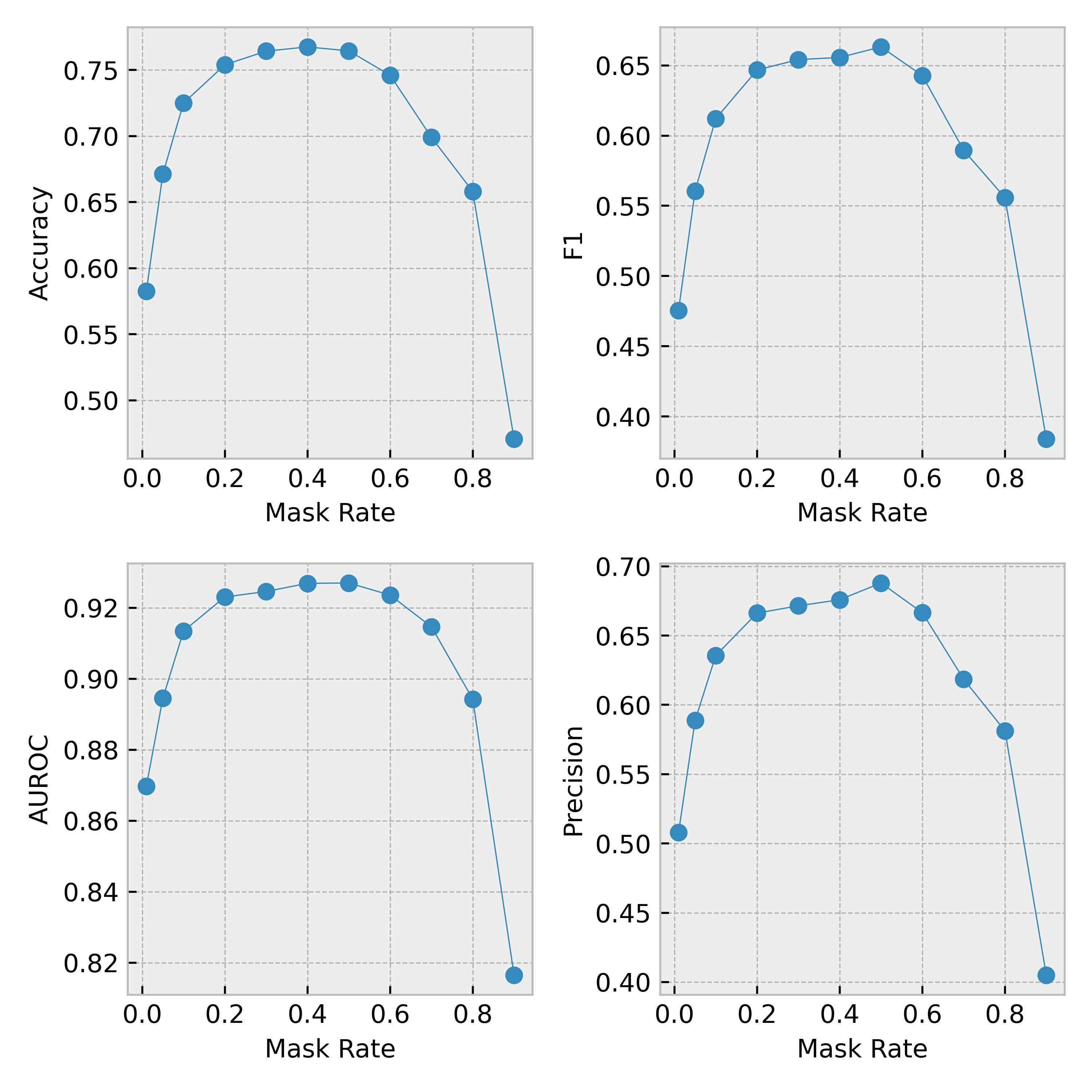}
  \caption{Test metrics i) accuracy ii) macro-averaged AUROC iii) macro-averaged F1 score and iv) macro-averaged precision, for the cancer classification task, against mask rate. Results shown for the \textbf{miRNA} datatset. Symbols indicate the test metric averaged over 5 seeds, each with different data splits and model weight initialisations. Solid lines added for visual aid. }
  \label{fig: optimal mask rate}
\end{figure}

To assess to what extent MTR contributed to the improvement in performance during pretraining, we compared the test metrics of the FTT, averaged over 5 different seeds, after pretraining with different levels of masking. The mask rate parameter of the FTT, $p_{m}\in[0,1]$
is the probability that any individual token is replaced with a mask token during pretraining. Masks are applied to each instance as they are passed through the model, such that at each epoch the data is masked differently, according to this random procedure. We pretrained an FTT for 200 epochs and then finetuned for a maximum of 200 epochs, using an early stopping based on the validation loss with a patience of 10 epochs. We computed the test metrics following this procedure, repeating this for 5 random seeds, corresponding to different data splits and model initialisations. In Figure \ref{fig: optimal mask rate} we find that the performance, across all metrics, improves as the mask rate increases, until reaching a peak around $p_{m} \approx 0.45$ at which point the performance decreases. The change in the test metrics is not symmetric about the peak, the performance decreases faster as $p_{m} > 0.5$ than when $p_{m}<0.5$, suggesting that MTR requires modest amounts of masking to provide an effective self-supervised signal. From this we infer that the model performs better for high levels of masking, such that the pretraining task is difficult enough to force the model to learn useful representations, whilst above an optimal value too much of the input is masked and the pretraining task becomes increasingly sub-optimal. 

\subsection{Performance of tabular deep learning against gradient boosting decision trees in the low-label regime}

GBDTs remain state of the art methods for classification and regression with tabular data, and it remains a point of research to determine when and how deep learning may outperform these classical methods, in the same way that they do in computer vision and NLP. GBDTs are fully supervised methods, and can not make use of unlabelled data and, hence, the low-label regime may be one such case where deep learning outperforms these methods. Two assess this we compared the performance of an FTT with and without pretraining against two popular GBDT models, XGBoost and CatBoost \citep{chen2016xgboost,prokhorenkova2018catboost} on the cancer classification task. For each of these models, we used their default hyperparameters. The pretraining phase of the FTT was set to last for 200 epochs. During finetuning the FTT was trained for a maximum of 200 epochs, with early stopping applied with 10 epochs of patience to the validation loss.

\begin{figure}[t]
  \centering
\includegraphics[width=0.95\linewidth]{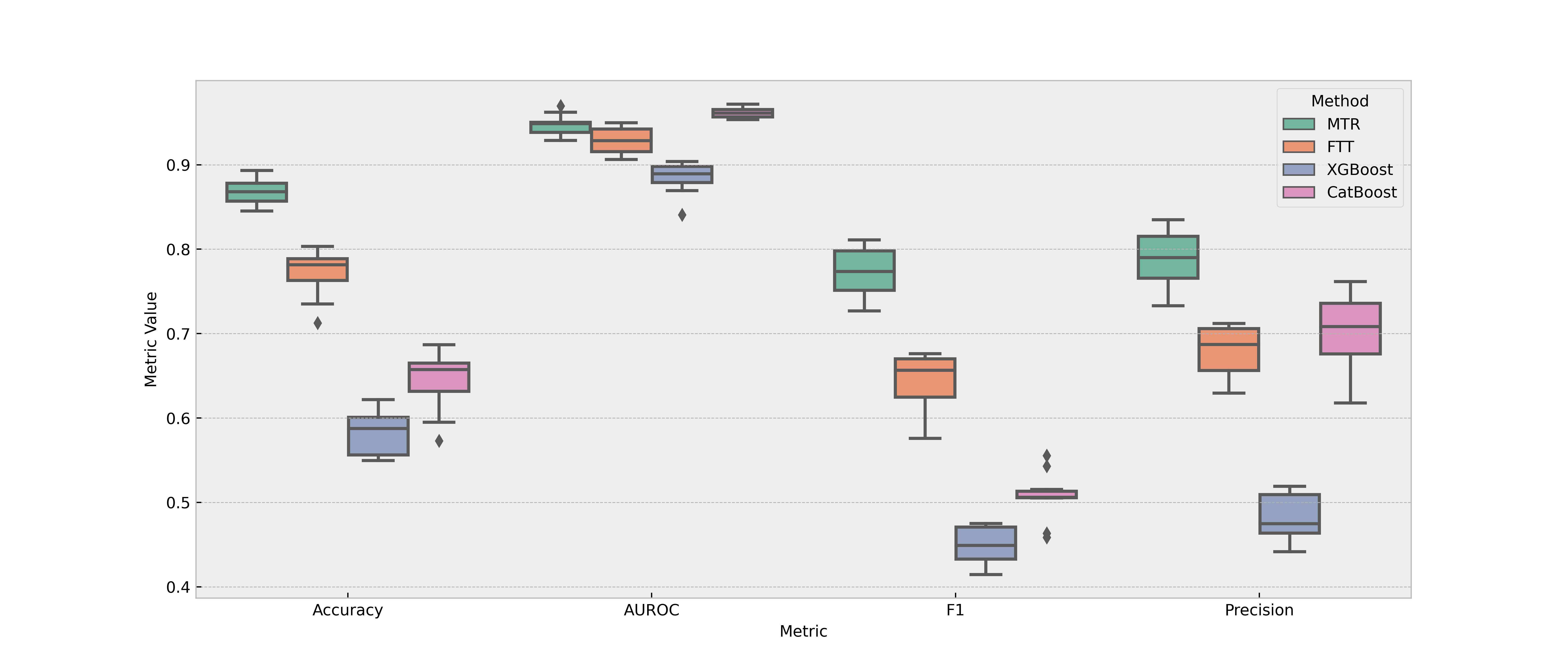}
  \caption{Performance on the cancer classification task, using the \textbf{mRNA} dataset, of 4 models: i) FTT ii) FTT pretrained with MTR, iii) CatBoost, and iv) XGBoost. In each case, default parameters for the model were used, and 1\% of the training set was used for supervised training. The test metrics shown are the overall accuracy, and macro-averaged AUROC, F1 score, and precision. The boxplots summarise the results of training the models across 10 different random seeds, corresponding to different data splits and weight initialisations. Outliers are represented by diamonds.}\label{fig: mrna gbdt}
\end{figure}

\begin{figure}
  \centering
\includegraphics[width=0.55\linewidth]{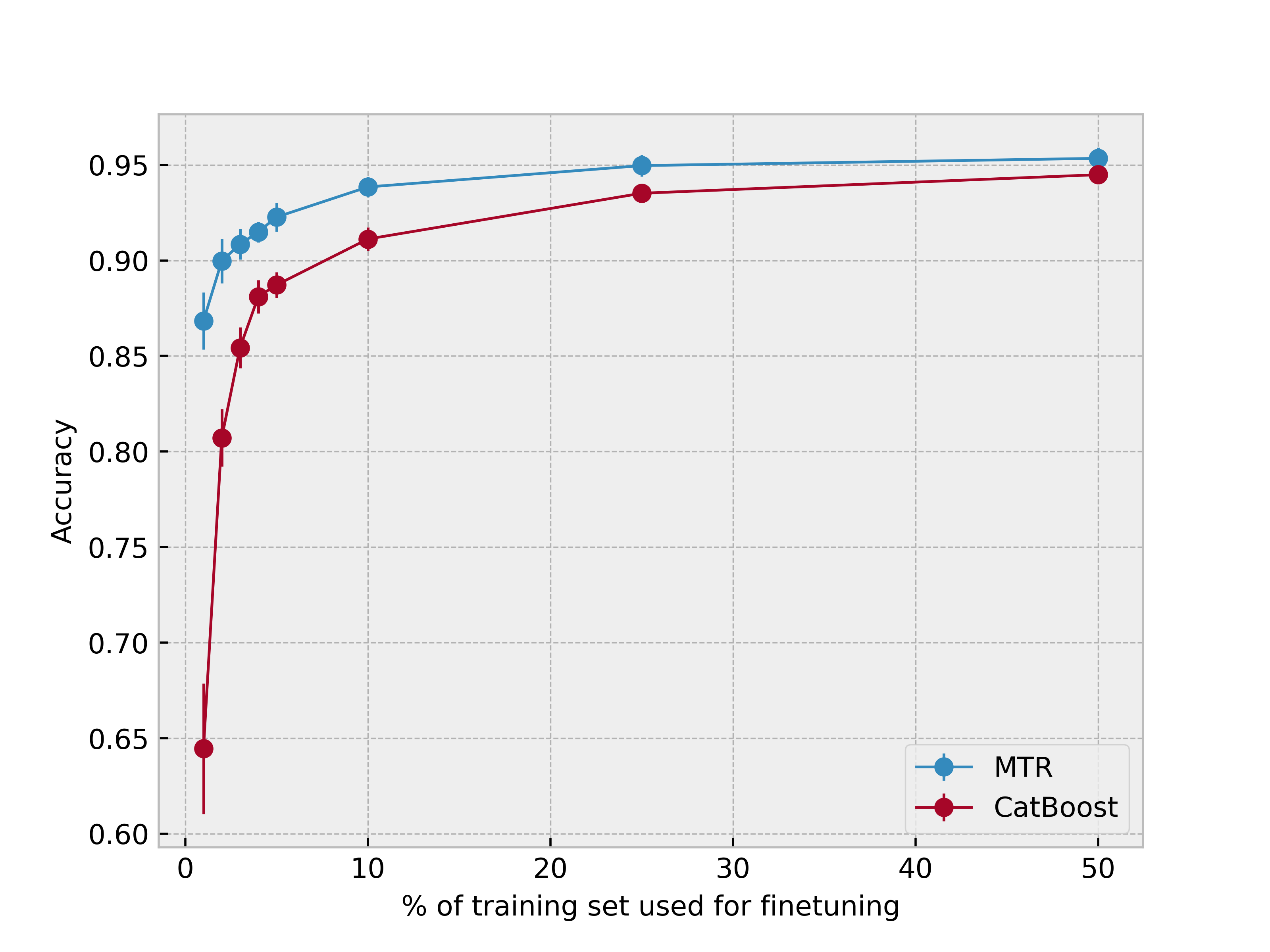}
  \caption{Overall accuracy on the cancer classification task, using the \textbf{mRNA} dataset, of a pretrained FTT and CatBoost, against the percentage of the training set that was used for supervised training. Symbols indicate average over 10 seeds corresponding to different data splits and model weight initialisation. Error bars correspond to standard deviation over 10 random seeds.}
  \label{fig: mrna accuracy vs labelled size}
\end{figure}

We show the results for the \textbf{mRNA} dataset, when 1\% of the training set was used for supervised training, in Figure \ref{fig: mrna gbdt}.  We find that the FTT pretrained via MTR outperforms all other models on overall accuracy, F1 score and precision. Additionally, the FTT without pretraining outperforms both GBDT methods in terms of overall accuracy, but not F1 and precision. These metrics are insensitive to class frequency, and hence tell us that the GBDT algorithm sacrifices overall accurate prediction across all classes. Indeed, it is well known that class imbalance poses a problem for training deep neural networks, and that learning from minority classes requires care. We repeated this experiment for both the \textbf{miRNA} and \textbf{RPPA} datasets. In Appendix \ref{app: further results} Table \ref{tab: GBDT comparison all omics} summarises the test metrics for each model, for each dataset. For all datasets, MTR performs significantly better than the FTT, CatBoost and XGBoost, in overall accuracy, F1 and precision. It is the only method that makes use of the unlabelled data in its training, and hence demonstrates that deep learning can be effective in the tabular domain when the amount of unlabelled data is much greater than the amount of labelled data. To further assess that this was the case, we compared the performance of the pretrained FTT and CatBoost (the stronger of the two GBDTs across our datasets) when different proportions of the training set is considered labelled data. In Figure \ref{fig: mrna accuracy vs labelled size} we show that the difference in performance between MTR and CatBoost decreases as the amount of training data used for supervised training is increased - supporting our claim that self-supervised pretraining allows deep models to outperform GBDTs in the low-label regime. However, in this instance MTR is significantly better than CatBoost until 50\% of the training data is used for finetuning. 

As part of our preprocessing, we applied PCA to each of our datasets, reducing each dataset to just 200 principle components. Since the dimensionality of the data was reduced, this preprocessing procedure incurred a loss of information. To assess whether this affected the relative performance of the FTT and GBDTs, we ran our experiments again without using PCA for dimensionality reduction (although we did not train the FTT on raw mRNA data, due to the high memory that the self-attention mechanism requires for datasets with thousands of features). The results are summarised in Appendix \ref{app: further results} in Table \ref{tab: raw GBDT comparison}. Across each of the datasets we find that the FTT, CatBoost and XGBoost each perform better without PCA included in the preprocessing. This is despite the fact that these models were trained on 1\% of the training set, where the number of labelled samples is significantly less than the number of features. For the \textbf{miRNA} and \textbf{RPPA} datasets, we find that CatBoost trained without PCA included in the preprocessing has superior performance to the FTT trained \textit{with} PCA, but that the FTT trained \textit{without} PCA is superior to CatBoost. For the \textbf{mRNA} dataset, however, CatBoost trained with or without PCA has weaker performance than the FTT trained with PCA. This implies that for datasets with a very high number of features relative to the number of labelled samples, the FTT is a viable model due to its ability to learn from very high-dimensional data reduced in dimensionality via PCA, hence avoiding having to train memory intensive models, whilst still out performing GBDTs trained with or without PCA. One additional observation is that CatBoost shows superior AUROC across each of the datasets, when trained on raw data, suggesting that the probabilities for incorrect classes are more widely distributed.

\begin{table}[t]
  \caption{Performance on the cancer classification task across \textbf{mRNA}, \textbf{miRNA} and \textbf{RPPA} datasets. We show results for 4 different models i) FTT ii) FTT pretrained with MTR, iii) CatBoost, and iv) XGBoost.  In each case, default parameters for the model were used, and 1\% of the training set was used for supervised training. The test metrics shown are the overall accuracy, and macro-averaged AUROC, F1 score, and precision. Test metrics are averaged over 10 seeds with (±) indicating the standard deviation. Best performance in \textbf{bold} and second best is \underline{underlined}.  }
  \label{tab: GBDT comparison all omics}
  \small
\begin{tabular}{lccccccl}\toprule
& \multicolumn{5}{c}{$1\%$ training data} 
\\\cmidrule(lr){3-6}
    &       & Accuracy & AUROC & F1   & Precision  \\ \midrule
 \textbf{mRNA}&FTT  & \underline{0.7712 ± 0.0281} &0.9286 ± 0.0157 &\underline{0.6413 ±0.0378}& 0.6796 ± 0.0310\\ 
 &MTR & \textbf{0.8684 ± 0.0157} & \underline{0.9469± 0.0128} & \textbf{0.7729 ±0.0295} & \textbf{ 0.7892±0.0322 } \\
 &CatBoost & 0.6445± 0.0360 &\textbf{ 0.9614±0.0061 } & 0.5066 ± 0.0298 & \underline{ 0.7016±0.0481 } \\
 &XGBoost &0.5828 ± 0.0272 & 0.8851 ± 0.0190 & 0.4490±0.0226 &0.4824 ±0.0274\\ \midrule
  \textbf{miRNA}&FTT  & \underline{0.6222± 0.0215} &0.8829 ±  0.0178&\underline{0.4918±0.0326} & 0.5272 ± 0.0375 \\
 &MTR & \textbf{ 0.7640± 0.0249} & \underline{0.9222 ±0.0158 } & \textbf{ 0.6584± 0.0317} & \textbf{ 0.6770± 0.0330} \\
 &CatBoost & 0.5652±0.0181  &\textbf{0.9372±0.0070} & 0.4403±0.0242  & \underline{0.6038±0.0319} \\
 &XGBoost  & 0.4812±0.0277  & 0.8322±0.0141 & 0.3628±0.0206  & 0.4020 ±0.0307 \\ \midrule
  \textbf{RPPA}&FTT  &\underline{0.5090±0.0276} & 0.8413±0.0203 &\underline{0.3865 ±0.0232} & 0.4209±0.0325 \\
 &MTR & \textbf{ 0.7297±0.0327 } & \textbf{0.9201 ± 0.0161} & \textbf{ 0.6201± 0.0411} & \textbf{0.6546 ± 0.0406} \\
 &CatBoost & 0.4183± 0.0455 &\underline{ 0.8691 ± 0.0204} & 0.2993 ± 0.0312 & \underline{ 0.4285±0.0569 } \\
 &XGBoost & 0.3635 ±  0.0363& 0.7799± 0.0213 &0.2607  ± 0.0295 & 0.2824± 0.0302\\
 \bottomrule
\end{tabular}
\end{table}

\begin{table}[ht]
  \caption{Performance on the cancer classification task across \textbf{mRNA}, \textbf{miRNA} and \textbf{RPPA} datasets. We show results for 4 different models i) FTT  ii) a second FTT iii) CatBoost, and iv) XGBoost, where we indicate models trained without PCA during preprocessing with an asterisk.  In each case, default parameters for the model were used, and 1\% of the training set was used for supervised training. The test metrics shown are the overall accuracy, and macro-averaged AUROC, F1 score, and precision. Test metrics are averaged over 10 seeds with (±) indicating the standard deviation. Best performance in \textbf{bold} and second best is \underline{underlined}. }
  \label{tab: raw GBDT comparison}
  \small
\begin{tabular}{lccccccl}\toprule
& \multicolumn{5}{c}{$1\%$ training data} 
\\\cmidrule(lr){3-6}
    &       & Accuracy & AUROC & F1   & Precision  \\ \midrule
 \textbf{mRNA}&FTT  & \textbf{0.7712 ± 0.0281} & \underline{0.9286 ± 0.0157} &\textbf{0.6413 ±0.0378}& \textbf{0.6796 ± 0.0310}\\
&FTT\textbf{*}&  -  & -  & - & - \\ 
 &CatBoost\textbf{*}& \underline{0.6743 ± 0.0267} & \textbf{0.9586 ± 0.0117} & \underline{0.5254 ± 0.0330} & \underline{0.6773 ± 0.0674} \\
  &XGBoost\textbf{*} & 0.5782 ± 0.0430& 0.8745 ± 0.0228  & 0.4547± 0.0350 & 0.4739 ± 0.0331 \\ \midrule
  \textbf{miRNA}&FTT  & 0.6222± 0.0215 &0.8829 ±  0.0178& 0.4918 ± 0.0326 & 0.5272 ± 0.0375 \\
  &FTT \textbf{*}&  \textbf{0.7242 ± 0.0188} & \underline{0.9166 ± 0.0144} & \textbf{0.6131 ± 0.0262} & \underline{0.6436 ± 0.0275} \\ 
  &CatBoost\textbf{*} & \underline{0.6342 ± 0.0302} & \textbf{0.9560 ± 0.0065} & \underline{0.5257± 0.0325} & \textbf{0.6500 ± 0.0537} \\
  &XGBoost\textbf{*} & 0.5761 ± 0.0226 & 0.9058± 0.0090& 0.4670 ± 0.0291 & 0.5155  ± 0.0310 \\ \midrule
  \textbf{RPPA}&FTT  & 0.5090 ± 0.0276 & 0.8413±0.0203 & 0.3865 ± 0.0232 & 0.4209 ± 0.0325 \\
  &FTT\textbf{*}& \textbf{0.6956 ± 0.0169} & \underline{0.9133 ± 0.0145} & \textbf{0.5623 ± 0.0180} & \textbf{0.5907 ± 0.0245 }\\ 
  &CatBoost\textbf{*}& \underline{0.5345 ± 0.0488}  & \textbf{0.9240 ± 0.0146} & \underline{0.4144 ± 0.0448} & \underline{0.5213 ± 0.0598} \\
  &XGBoost\textbf{*} & 0.4808 ± 0.0379  & 0.8664 ± 0.0228 & 0.3773± 0.0348 & 0.4039 ± 0.0377\\
 \bottomrule
\end{tabular}
\end{table}

\subsection{Tabular transformers perform well against benchmarks after hyperparameter optimisation}\label{sec: hpo}

A caveat to our results showing that the FTT is a more suitable model for tabular omics data than GBDTs in the low-label regime, is that so far we have used default parameters for each of the models we have considered. It may have been that after hyperparameter optimisation, the qualitative nature of our results change. However, in many practical situations, particularly in the low-label regime, hyperparameter optimisation (HPO) may not be suitable, since sufficient labels are needed to form a validation set in order for HPO to yield sufficient generalisation. Hence, when considering the low-label regime, the reporting of results with default parameters is crucial. However, to further assess the suitability of the FTT and pretraining via MTR for tabular omics data from TCGA, we investigated the effects of HPO. 

We compared the FTT with and without pretraining to CatBoost and XGBoost after HPO, to ensure that their success was not dependent on default parameters alone. We also compared these models to two additional benchmarks, namely, 
\begin{itemize}
\item \textbf{MLP}: a simple multi-layer perceptron. We included this benchmark as it allowed us to assess whether it is necessary to have a complex architecture like the FTT for tabular data problems. 
\item \textbf{Value Imputation and Mask Estimation (VIME)}:  a self-supervised framework that pioneered self-supervised learning for tabular data \citep{yoon2020vime}. We restrict this model, using the encoder trained via self-supervised learning with several MLP layers attached as a predictive head, and ignore the additional semi-supervised training objectives recommended by the authors, so that we may compare the effect of its reconstruction based self-supervised objective to our contrastive one.  
\end{itemize}
In Table \ref{tab: hpo benchmarks} we compare the test metrics of each of the models over 5 seeds, after HPO. The details of the parameters tuned in HPO can be found in Appendix \ref{app: hpo}. We find that across each dataset, either the MLP, FTT or both, outperform both GBDT methods in terms of overall accuracy and F1 score. CatBoost is found to outperform XGBoost on every metric across each dataset, and also achieves the best precision for the mRNA and miRNA datasets, out of the models which do not utilise unlabelled data. While these results still suggest that there is no universal solution to whether deep learning or classical methods should be applied to tabular data problems, it does suggest that deep learning is particular effective for datasets with low numbers of samples and high dimensionality. This is particularly true for datasets where the number of features greatly outnumbers the number of labelled samples, where the FTT is able to learn effectively from a dimensionality-reduced dataset. Comparing the FTT and MLP, we find that the MLP outperforms the FTT on every metric for the \textbf{miRNA} and \textbf{RPPA} datasets. The FTT, however, outperforms the MLP on overall accuracy, F1 and precision for the \textbf{mRNA} dataset. Hence, for datasets with a very large number of features, orders of magnitude larger than the number of labelled samples, the FTT is the preferred model. 

We find that across each dataset, the model which performs best on accuracy, F1 and precision, is one of the models that utilises self-supervised pretraining. However, the MLP has the highest AUROC across all datasets, suggesting that while it is less accurate, it narrows down its decision between fewer classes, rejecting more incorrect classes. Comparing the two self-supervised methods, we find that MTR outperforms VIME on all metrics for the \textbf{mRNA} and \textbf{RPPA} datasets. For the \textbf{miRNA} datasets, MTR outperforms VIME on overall accuracy and F1. Furthermore, for the \textbf{mRNA} dataset VIME performs worse on all metrics than all other deep learning methods, even those which are not pretrained. This suggests that the self-supervised pretraining VIME utilises is ineffective for datasets with a very large number of features, especially in the low-label regime. This suggests that a reconstruction based self-supervised objective requires significantly more data than a contrastive objective, when the number of features is high.

\begin{table}[t]
  \caption{Performance of several models on the cancer classification task, across different datasets, after hyperparameter optimisation. Details of hyperparameter optimisation is described in Appendix \ref{app: hpo}. The test metrics shown are the overall accuracy, and macro-averaged AUROC, F1 score, and precision. Test metrics are averaged over 5 seeds, with different data splits, weight initialisations, and HPO, with (±) indicating the standard deviation across seeds. Best performance in \textbf{bold} and second best is \underline{underlined}. }
  \label{tab: hpo benchmarks}
  \small
\begin{tabular}{lccccccl}\toprule
& \multicolumn{5}{c}{$1\%$ training data} 
\\\cmidrule(lr){3-6}
    &       & Accuracy & AUROC & F1   & Precision  \\ \midrule
 \textbf{mRNA}&FTT   & \underline{0.8177 ± 0.0241}  & 0.9389 ± 0.0175 & \underline{0.7045 ± 0.0553} & 0.7340 ± 0.0528 \\ 
& MLP &  0.7984 ± 0.0338  & \textbf{0.9758 ± 0.0067}  & 0.6781 ± 0.0482 & 0.7026 ± 0.0479 \\ 
& MTR &  \textbf{0.8854 ± 0.0143}  & 0.9485 ± 0.0146  & \textbf{0.7896 ± 0.0278} & \textbf{0.8007 ± 0.0336} \\ 
& VIME &  0.6368 ± 0.0691  &  0.9306 ± 0.0168 & 0.5587 ± 0.0567 & 0.6848 ± 0.0522 \\ 
& CatBoost &  0.7407 ± 0.0389  & \underline{0.9725 ± 0.0048}  & 0.6443 ± 0.0509 & \underline{0.7707 ± 0.0157} \\ 
& XGBoost &  0.6112 ± 0.0354  & 0.9121 ± 0.0200  & 0.5148 ± 0.0328 & 0.5589 ± 0.0477 \\  \midrule
  \textbf{miRNA}&FTT   & 0.6590 ± 0.0289  & 0.9119 ± 0.0132  & 0.5284 ± 0.0324 & 0.5688 ± 0.0356 \\ 
& MLP &  0.7143 ± 0.0279  & \textbf{0.9627 ± 0.0063}  & 0.5939 ± 0.0142 & 0.6232 ± 0.0124 \\ 
& MTR &  \textbf{0.7510 ± 0.0323 }& 0.9304 ± 0.0160  & \textbf{0.6503 ± 0.0435} & 0.6731 ± 0.0499 \\ 
& VIME & \underline{0.7126 ± 0.0295} & \underline{0.9420 ± 0.0228} & \underline{0.6453 ± 0.0544 } & \textbf{0.7152 ± 0.0416}   \\ 
& CatBoost &  0.6250 ± 0.0148  & 0.9421 ± 0.0095  & 0.5404 ± 0.0296 & \underline{0.6777 ± 0.0502} \\ 
& XGBoost &  0.5125 ± 0.0109 & 0.8585 ± 0.0016  & 0.4340 ± 0.0144 & 0.4772 ± 0.0254 \\  \midrule
  \textbf{RPPA}&FTT   &  0.5307 ± 0.0247  & 0.8574 ± 0.0223  & 0.3818 ± 0.0355 & 0.4199 ± 0.0407 \\ 
& MLP &  0.6516 ± 0.0164  & \textbf{0.9455 ± 0.0036}  & 0.5290 ± 0.0292 & 0.5604 ± 0.0305 \\ 
& MTR &  \textbf{0.7824 ± 0.0331} & \underline{0.9386 ± 0.0163}  &\textbf{ 0.6640 ± 0.0353} & \textbf{0.6952 ± 0.0300 }\\ 
& VIME & \underline{ 0.6705 ± 0.0337 } & 0.9340 ± 0.0071  & \underline{0.5669 ± 0.0476} & \underline{0.6124 ± 0.0576} \\ 
& CatBoost &  0.5085 ± 0.0526  & 0.8869 ± 0.0281  & 0.3949 ± 0.0543 & 0.5409 ± 0.0981 \\ 
& XGBoost &  0.3938 ± 0.0281  & 0.7892 ± 0.0351  & 0.2916 ± 0.0319 & 0.3335 ± 0.0693 \\  \midrule
 \bottomrule
\end{tabular}
\end{table}

\subsection{Mask Token Replacement for handling missing features}

We have demonstrated that MTR is effective as an augmentation in contrastive self-supervised learning, when applied to tabular expression data. Earlier work has also shown that it is an effective augmentation for supervised training \citep{onishi2023rethinking}. In effect, MTR replaces features at random with a value which is a learned parameter of the model.  Aside from augmentation during training, another potential, but as yet untested, application of this learned imputation is in the handling of missing features. For datasets with samples that have features that are corrupted or otherwise missing, with the set of features that are missing varying between each incomplete sample, imputation can allow models to learn and make predictions from incomplete samples. It was our hypothesis that MTR, as a learned imputation, should be effective when handling missing features. As the FTT is trained with data augmented by MTR, the model learns from samples whose features have been masked at random - and hence should be robust when making predictions from new samples which are truly incomplete. 

To test this hypothesis we trained several models on the \textbf{RPPA} dataset. After splitting our dataset into train/validation/test sets as before, we standardised our data, based on the mean and variance of the training set, and did not apply PCA. We then created a test set with incomplete samples, synthetically. Each sample in the test set was selected with probability $p_{\mathcal{I}}$ to be incomplete. For each incomplete sample, each feature was selected with probability $p_{\mathcal{M}}$ to be missing. By training models on the complete \textbf{RPPA} dataset, and testing on samples with missing data, we could determine how robust a model was to the distribution shift incurred by features missing at random in the test set. 

We compared four models 
\begin{enumerate}
\item FTT trained without augmentation: at test time missing features are replaced with the mean value of that feature.
\item FTT trained with MTR as augmentation, with mask rate equal to the rate of missing features $p_{m}=p_{\mathcal{M}}$. At test time missing features are replaced with mask token learned during training. 
\item FTT pretrained via MTR, and then finetuned without augmentation, at test time missing features are replaced with mask token learned during training. 
\item CatBoost with default parameters. At test time the default missing features handler is used, imputing missing features with minimum value of that feature.  
\end{enumerate}

\begin{figure}[t]
  \centering
\includegraphics[width=0.45\linewidth]{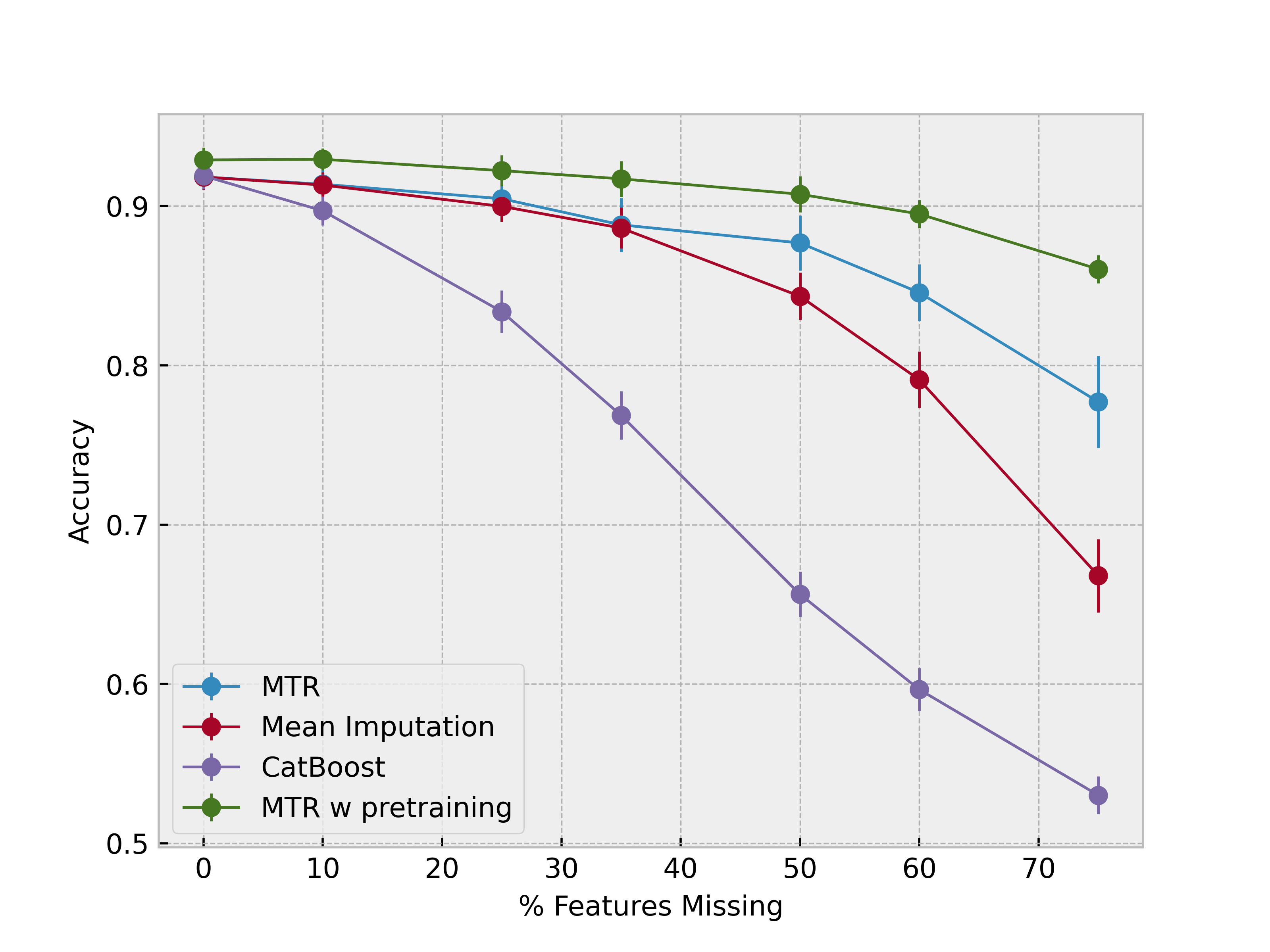}
\includegraphics[width=0.45\linewidth]{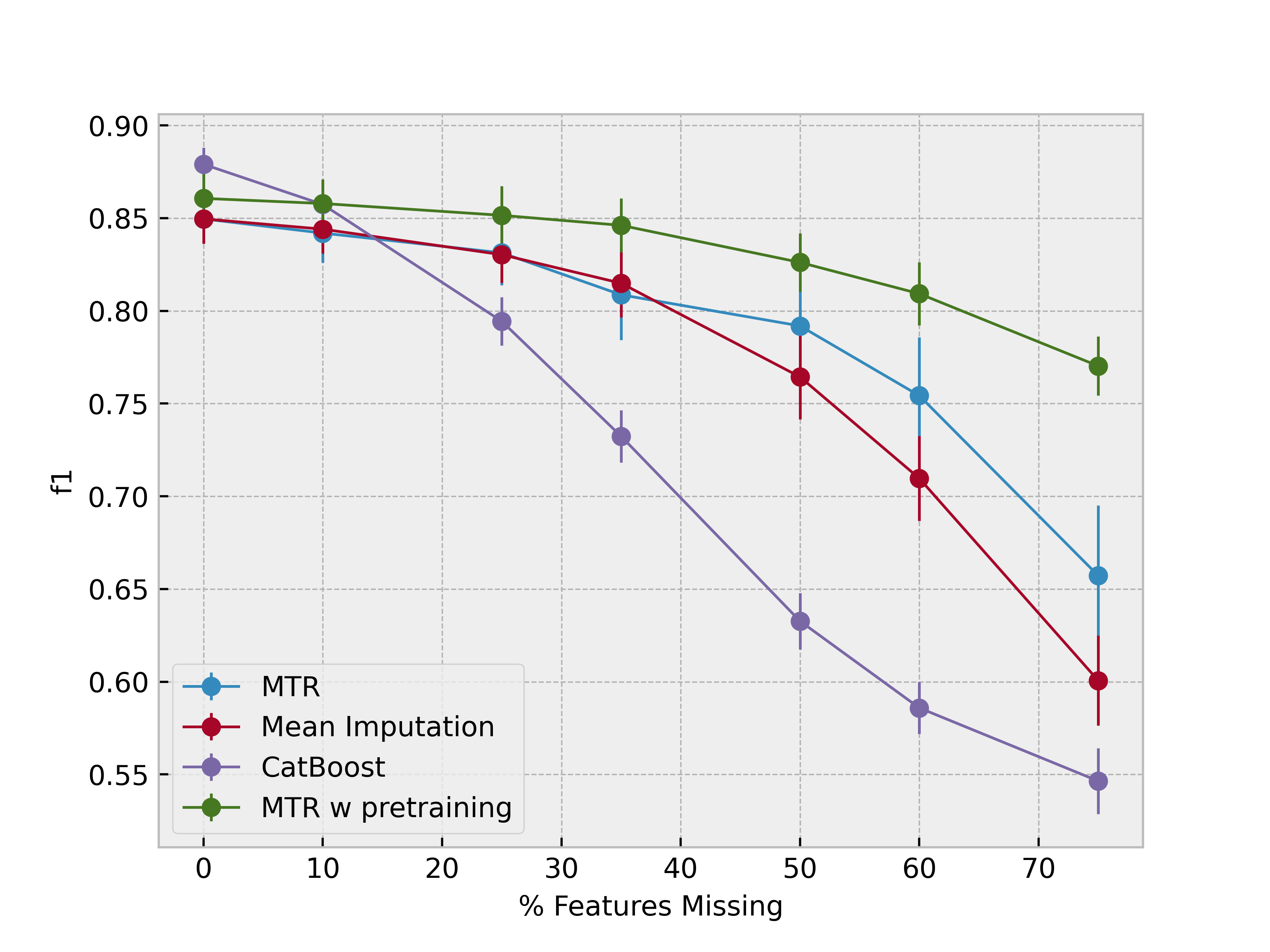}
  \caption{Test accuracy (left) and macro averaged F1 score (right)  against average \% of features missing in an incomplete sample $p_{\mathcal{M}}$.  Samples in the test set were selected to be incomplete with probability $p_{\mathcal{I}}=0.5$. Each figure shows results for four models i) FTT trained without augmentation, with mean imputation at test time, ii) FTT trained with MTR as augmentation with mask token imputing missing values at test time, iii) FTT pretrained via MTR, finetuned with MTR as augmentation with mask token imputing missing values at test time, iv) CatBoost with default missing features handling. Symbols indicate the mean, and error bars show standard deviation, over 10 seeds, which set the data splits, weight initialisation and test set construction.  }
  \label{fig: rppa accuracy vs fraction missing}
\end{figure}

We constructed our test set to have a high fraction of incomplete samples, setting $p_{\mathcal{I}}=0.5$ such that approximately half of samples were incomplete. In Figure \ref{fig: rppa accuracy vs fraction missing} we show the accuracy and macro averaged F1 score against the average fraction of features which were missing from incomplete samples, $100 \times p_{\mathcal{M}}$. Each model was trained using 10\% of the training set as a labelled dataset - in this region we find that CatBoost outperforms the FTT when all samples are complete, but with pretraining the FTT is superior in terms of overall accuracy. As the fraction of features missing from incomplete samples increases, each model is found to decrease in accuracy and F1 score.  However, all models are shown to be robust to missing features - with 50\% of features missing the least accurate model, CatBoost, has an overall accuracy greater than 75\%, such that it still correctly classifies at least half of the samples which are missing half of the features the model is trained on. CatBoost degrades significantly with the fraction of features removed, with a significant difference between all other models shown with just 20\% of features removed. We find that imputing the missing features with the learned mask token is more effective than mean imputation, with a significant difference in performance between these two methods once 50\% of features are missing. The difference in performance between these two methods increases as the fraction of missing features increases, with MTR showing the best performance in the extreme case when 75\% of features are missing. We also find that contrastive pretraining via MTR makes the FTT even more robust to missing features at test time, performing the best for all levels of missing features in overall accuracy, and macro-averaged F1 when more than 25\% of features are missing from incomplete samples.  

\begin{figure}[t]
  \centering
\includegraphics[width=0.45\linewidth]{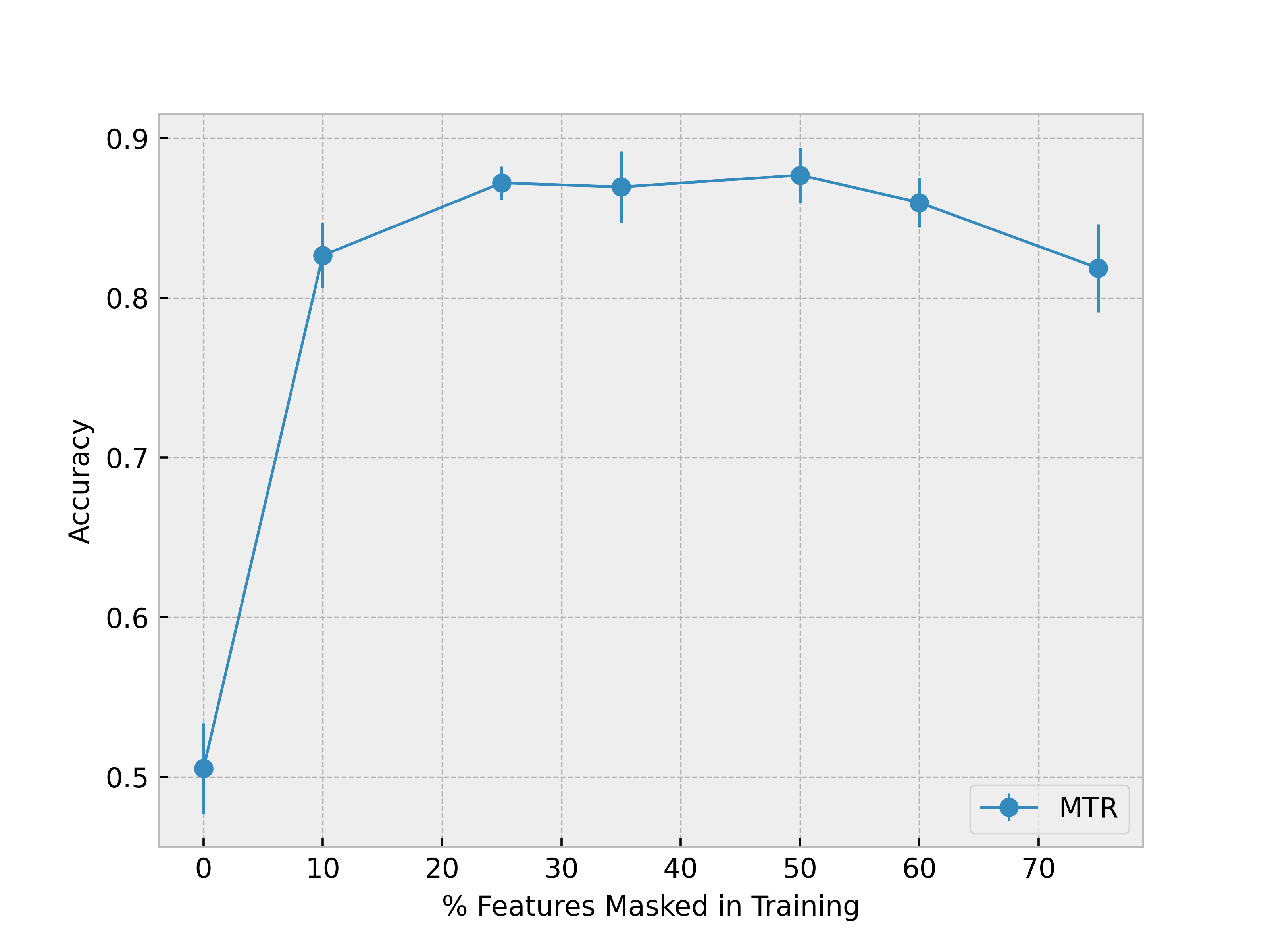}
  \caption{Test accuracy of a FTT which imputes missing features with mask token at test time, against mask rate used during training with MTR as augmentation. Samples in the test set were selected to be incomplete with probability $p_{\mathcal{I}}=0.5$, with each feature selected to be missing with probability $p_{\mathcal{M}}=0.5$, Symbols indicate the mean, and error bars show standard deviation, over 10 seeds, which set the data splits, weight initialisation and test set construction.  }
  \label{fig: missing rppa accuracy vs mask}
\end{figure}

To assess the impact that MTR during training has on robustness to missing features at test time, we constructed a test set with $p_{\mathcal{I}} = p_{\mathcal{M}}=0.5$, such that half the samples were incomplete and each of them had half of their features missing. We then trained several FTTs, each with different mask rate $p_m$ during training. As shown in Figure \ref{fig: missing rppa accuracy vs mask} if the model is trained without MTR as augmentation, the performance of the model is poor. With low levels of masking during training, the performance of the model has higher accuracy, and the accuracy peaks between 25-60\% of the features being masked during training.  Model performance begins to decrease for masking that is significantly higher than the fraction of missing features seen at test time. 

This tells us that masking during training allows the model to become robust to the distribution shift incurred due to missing features in the test set. This is true even if significantly more or less features are missing at test time than were masked during training. The robustness to the level of features masked during training suggests that the model learns to ignore the presence of mask tokens, via self-attention, and learn from features which are unmasked. This is a significant advantage of MTR as a method for handling missing features, as it is generally not known how many features will be missing at test time, and samples may vary greatly in the number of features which are missing.

\section{Multi-modal masked transformers for multi-omics integration }

\setcounter{figure}{0}
\renewcommand{\thefigure}{\arabic{section}.\arabic{figure}}

\setcounter{table}{0}
\renewcommand{\thetable}{\arabic{section}.\arabic{table}}

\begin{figure}
  \centering
\includegraphics[width=0.65\linewidth]{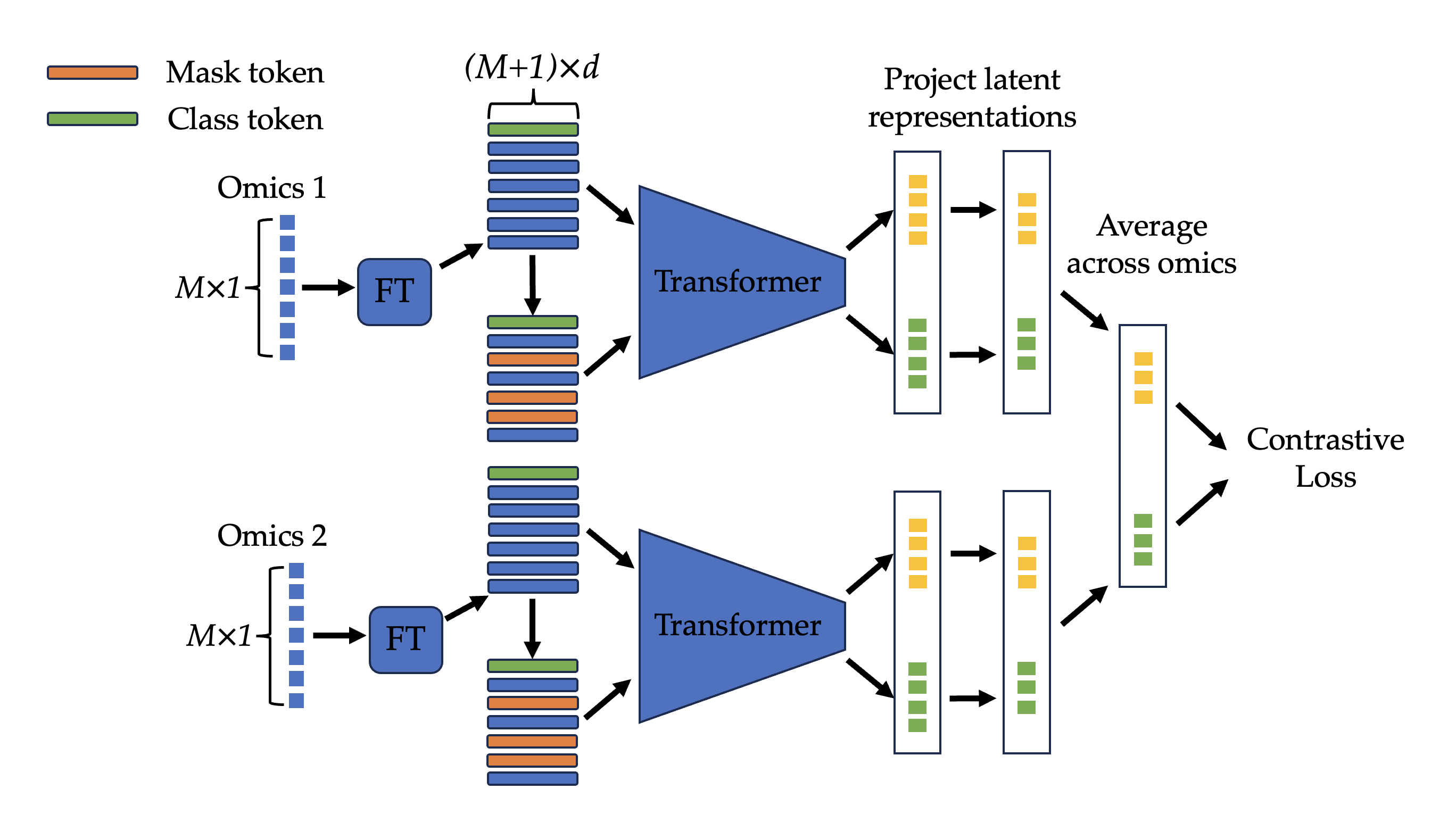}
  \caption{Sketch of DuoFTT architecture during contrastive pretraining. Each omics, here shown to have $M$ features, is tokenised, such that they are transformed to $d$-dimensional vectors. The stack of $M$ tokens, with an additional class token, from each omics, is passed through an omics-specific transformer, which produces a latent representation. The stack of $M$ tokens is then masked, such that a random fraction of tokens is replaced with the mask token. The new stack of tokens, from each omics, is passed through the omics-specific transformers, which produces latent representations corresponding to the masked inputs. The latent representations from each omics-specific transformer are then projected via omics-specific linear layers. Finally, the projections are averaged over the 2 omics, and the two latent representations, corresponding to the original and masked inputs, are aligned via the contrastive loss function. }
  \label{fig: duo ftt}
\end{figure}

After establishing that contrastive learning with MTR is an effective pretraining scheme for tabular omics data, we now focus on whether it can be used for multi-omics integration. We propose a novel model for multi-omics integration, which we name the DuoFTT, which we illustrate in Figure \ref{fig: duo ftt}. This model is modular by design, it takes data from two omics at input, with each omics passed through their own FTT. The class token from each FTT is projected via linear layers, and the projection is then averaged across the two omics. Just as is the case with the unimodal FTT described in the previous section, during pretraining, a masked and unmasked version of the instance of data is passed through the DuoFTT, yielding two projected latent representations. To create the masked instance of data, features from both omics are chosen with probability $p_{m}$ to be replaced with the mask token. The model is then trained minimising the NTXent loss between the two projected latent representations. This multi-modal model is an example of a late-fusion model as the features from each modality are fused (via the element-wise averaging of the projected latent representations across omics) in latent space. A downside of this is that it does not explicitly model the correlation between features from different modalities - however we found that a late-fusion model that is modular by design has a number of advantages, which we later comment on.

 During supervised finetuning, the class token from each omics-specific FTT is passed through omics-specific linear layers, and an omics-specific output layer for multi-class classification. The output layers from each omics are element-wise averaged, and this is the final output of the DuoFTT which is then passed to a cross entropy loss function.

\subsection{Two omics are better than one}

The DuoFTT is the first instance of creating a multi-modal model which is pretrained with contrastive learning and MTR. To evaluate whether this architecture and pretraining scheme is effective for multi-omics data in the low-label regime, we compared the model with and without pretraining. To assess whether contrastive learning with MTR effectively integrates data from different omics, we also compared the performance of the DuoFTT with two FTTs, pretrained and fine-tuned on the same samples, but with features from one of the two omics sources with which the DuoFTT was trained.

We trained these models using data from each of the \textbf{miRNA+RPPA}, \textbf{mRNA + RPPA} and \textbf{mRNA + miRNA} subsets of the TCGA. In these datasets only patients where both omics are measured were considered. In each case, we split 20\% of the dataset into a test set. The remaining data was split 90:10 into train and validation data. 1\% of the training set was used for finetuning. All splits were stratified such that all splits had the same class imbalance. We pretrained DuoFTT and the FTTs for 200 epochs. All models were finetuned for a maximum of 200 epochs, with early stopping based on the validation loss, with a patience of 10 epochs. 

\begin{figure}
  \centering
\includegraphics[width=0.55\linewidth]{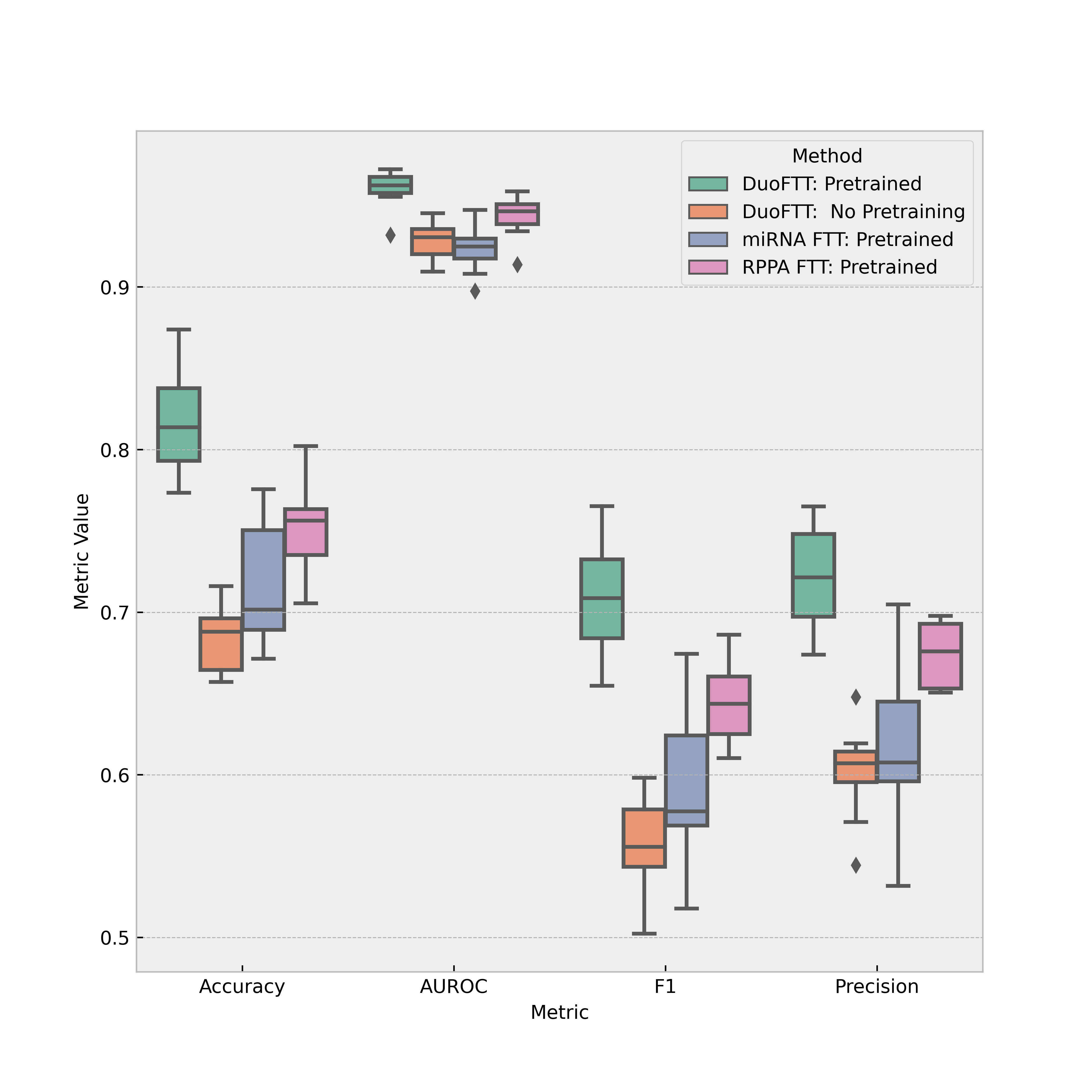}
  \caption{Test metrics of 4 models: i) DuoFTT finetuned on miRNA and RPPA, ii) DuoFTT also pretrained with MTR, iii) FTT pretrained and finetuned with miRNA and, iv)  FTT pretrained and finetuned with RPPA. The test metrics shown are the overall accuracy, and macro-averaged AUROC, F1 score, and precision. Results are shown for 10 random seeds with different data splits and weight initialisations. Outliers are represented by diamonds.}
\label{fig: joint mtr}
\end{figure}

In Figure \ref{fig: joint mtr} we show results for models trained on the \textbf{miRNA+RPPA} dataset and summarise the results of this experiment across each dataset in Table \ref{tab: duoFTT}. When comparing the DuoFTT with and without pretraining, we see that MTR provides an effective self-supervised signal, yielding drastic improvements in overall accuracy, and significant improvements across all metrics. This is consistent across all datasets. For the \textbf{miRNA+RPPA}, \textbf{mRNA + RPPA} datasets, the pretrained DuoFTT is the best performing model across all test metrics. For the \textbf{mRNA + miRNA} dataset, the pretrained DuoFTT and the pretrained FTT with mRNA data are the best performing models, with no significant difference between them. 
If we compare the DuoFTT without pretraining, we see that in some cases this model outperforms the FTTs pretrained and finetuned on a single omics. Hence, for some combination of omics the benefit of combining multiple omics into one predictive model can outweigh the benefit of self-supervised pretraining on single omics measurements. From this we conclude that multi-omics integration and self-supervision are complementary and in tandem provide the best performance, but in cases where unlabelled data is available in one omics but not other, the benefit of multi-omics integration is less clear. 

\begin{table}[t]
  \caption{For each dataset, with measurements from 2 different omics, we compare 4 models: i) DuoFTT finetuned on both omics ii) DuoFTT also pretrained with MTR on both omics, iii) an FTT pretrained and finetuned on a single omics and iv) an FTT pretrained and finetuned on the other omics.  The test metrics shown are the overall accuracy, and macro-averaged AUROC, F1 score, and precision.Test metrics are averaged over 10 seeds with (±) indicating the standard deviation. Best performance in \textbf{bold} and second best is \underline{underlined}.  }
  \label{tab: duoFTT}
 \small
\begin{tabular}{lccccccl}\toprule
& \multicolumn{5}{c}{$1\%$ training data} 
\\\cmidrule(lr){3-6}
    &       & Accuracy & AUROC & F1   & Precision  \\ \midrule
 \textbf{miRNA+RPPA}& DuoFTT  & 0.6838 ± 0.0204 &0.9286 ± 0.0122 & 0.5566 ±0.0294& 0.6013± 0.0280\\ 
 & DuoFTT+MTR & \textbf{0.8177 ± 0.0316} & \textbf{0.9605±  0.0114} & \textbf{0.7062 ±0.0360} & \textbf{ 0.7225±0.0303 } \\
 & miRNA only & 0.7159± 0.0390 & 0.9232±0.0137 &0.5961 ±0.0489& 0.6186±0.0503  \\
 & RPPA only & \underline{0.7546 ± 0.0295} & \underline{0.9433 ± 0.0126} & \underline{0.6460±0.0257} &\underline{0.6741 ±0.0208} \\ \midrule
  \textbf{mRNA+miRNA}& DuoFTT  &0.7823  ±0.0264 &0.9546 ±0.0075  & 0.6759±0.0365& 0.7312± 0.0394 \\ 
 & DuoFTT+MTR & \underline{0.8859± 0.0139}& \textbf{0.9702±0.0100}   & \underline{0.8180 ± 0.0284}& \underline{0.8377± 0.0269}\\
 & mRNA only & \textbf{0.8895±0.0238} & \underline{0.9682±0.0096} & \textbf{0.8349±0.0366}&\textbf{0.8522 ±0.0359}\\
 & miRNA only & 0.7696±0.0249  & 0.9429 ± 0.0087  & 0.6755± 0.0386& 0.7040±0.0413 \\ \midrule
  \textbf{mRNA+RPPA}& DuoFTT  & 0.7994 ± 0.0211& 0.9600± 0.0105 & 0.7219± 0.0152& 0.7666± 0.0161 \\ 
 & DuoFTT+MTR & \textbf{0.9237± 0.0158} & \textbf{0.9787± 0.0106} & \textbf{0.8706 ± 0.0238}  & \textbf{0.8831± 0.0300} \\
 & mRNA only & \underline{0.8802 ± 0.0219} & \underline{0.9754± 0.0087}& \underline{0.8199± 0.0297}& \underline{0.8318±0.0317} \\
 & RPPA only & 0.7215± 0.0593 & 0.9339 ± 0.0240  & 0.6333 ± 0.0629 & 0.6747± 0.0633\\ 
 \bottomrule
\end{tabular}
\end{table}

\subsection{Masked transformers are effective cross-omics learners}

Multi-modal deep learning models can be designed such that each modality is processed by a different module that can be extracted and used for predictions with data from that modality alone. A potential benefit of such an approach is that by training a multi-modal model, the representations learned by each module may be stronger and better suited for downstream tasks. Hence, it is possible that by training a multi-modal model in a self-supervised manner, and then extracting the modality-specific module and finetuning it for a particular downstream task, that the performance of this model may be greater than a model pretrained and finetuned on data from that modality alone. In other words, multi-modal pretraining may yield stronger unimodal predictions. This was recently demonstrated with a multi-modal model that considered images and tabular data \citep{hager2023best}. The DuoFTT is designed to be separable, with a FTT handling each omics separately. It is pretrained, however, in a joint manner - the projected latent representations of each omics-specific FTT are averaged and it is this output that is passed to the contrastive objective function. Hence, we investigated whether a pretrained DuoFTT can be used to train FTTs that make better predictions from a single omics source. 

For the \textbf{miRNA+RPPA}, \textbf{mRNA + RPPA} and \textbf{mRNA + miRNA}  datasets, we pretrained a DuoFTT using MTR. After pretraining, the omics-specific FTTs were extracted and finetuned on their respective omics. For comparison, we pretrained and finetuned 2 FTTs, one for each of the omics in the dataset considered. Hence, for each dataset, we trained 4 FTTs in total, 2 for each omics represented in the dataset. For each omics in the dataset, one FTT would be pretrained with multi-modal data (by extracting it from a pretrained DuoFTT) and the other would be pretrained using the same samples of data but with just the features from omics. Both FTTs would be finetuned using the same single omics features. Crucially,  to keep model comparison fair, each FTT was trained using the same samples, only the features they were pretrained with would differ between the models. All models were pretrained for 200 epochs, and finetuned on $1\%$ of the training set for a maximum of 200 epochs, using validation loss as an early stopping criteria with a patience parameter of 10 epochs. 

\begin{figure}[t]
  \centering
\includegraphics[width=0.55\linewidth]{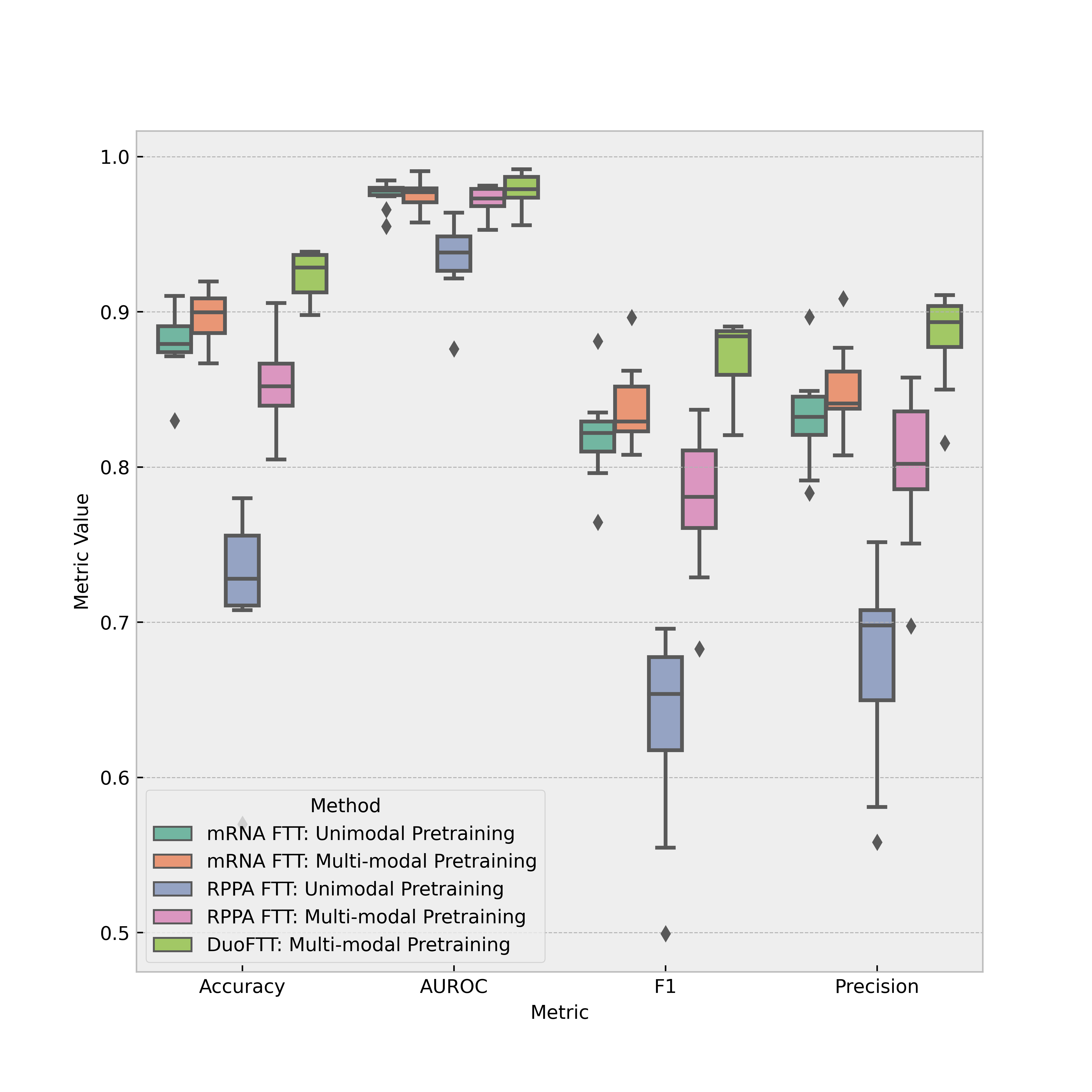}
  \caption{Test metrics of 5 models: i) DuoFTT pretrained and finetuned on miRNA and RPPA, ii) FTT pretrained with miRNA and RPPA, finetuned on RPPA, iii)  FTT pretrained with miRNA and RPPA, finetuned on miRNA, iv) FTT pretrained and finetuned with miRNA, v) FTT pretrained and finetuned with RPPA. The test metrics shown are the overall accuracy, and macro-averaged AUROC, F1 score, and precision. Results are shown for 10 random seeds with different data splits and weight initialisations. Outliers are represented by diamonds. }
 \label{fig: cross omics}
\end{figure}

In Figure \ref{fig: cross omics} we find that a FTT pretrained on both mRNA and RPPA, and then finetuned using RPPA only, performed better than a FTT pretrained and finetuned on the same RPPA data, without multi-modal pretraining. We refer to this phenomenon as `cross-omics learning', as the model has been pretrained with mRNA features, but these features were not passed through the FTT during pretraining. During pretraining, the model has associated mRNA features with RPPA features, and this has led to stronger representations. Similarly, we also find in Figure \ref{fig: cross omics} that the FTT finetuned on mRNA data benefits from multi-modal pretraining, when compared with the same FTT pretrained with mRNA features only, but the improvement in this case is slight. Hence, this demonstrates that multi-omics pretraining can create better unimodal models - but that the benefit is omics specific. Indeed in Table \ref{tab: cross omics summary}, we find that cross-omics learning is only present in a few cases. 

From Table \ref{tab: duoFTT}  we find that the performance of the DuoFTT pretrained \textit{and} finetuned on both miRNA and RPPA is significantly better than any of the single omics models, with or without multi-modal pretraining. This shows that both mRNA and RPPA contain useful modality-specific information for the cancer classification task. Despite this, pretraining with miRNA and RPPA leads to worse unimodal predictions with miRNA alone. We hypothesise that this is due to the relative importance of the features of miRNA and RPPA for this particular task. By pretraining on both miRNA and RPPA in a joint manner, all miRNA and RPPA features are put on equal footing in the self-supervised task - contrastive learning with MTR trains the model to provide similar latent representations given a random set of input features being masked out.

\begin{table}[t]
  \caption{For each dataset we compare two FTTs: one that has been separated from a DuoFTT pretrained on both omics, and finetuned on one omics, and another which is pretrained and finetuned on a single omics only. The test metrics shown are the overall accuracy, and macro-averaged AUROC, F1 score, and precision. Test metrics are averaged over 10 seeds with (±) indicating the standard deviation. Best performance in \textbf{bold}.  }
  \label{tab: cross omics summary}
  \small
\begin{tabular}{lccccccl}\toprule
& \multicolumn{5}{c}{$1\%$ training data} 
\\\cmidrule(lr){3-6}
    &       & Accuracy & AUROC & F1   & Precision  \\ \midrule
 \textbf{miRNA + RPPA}& miRNA: unimodal  &  \textbf{0.7159±0.0390} & \textbf{0.9232±0.0137}  & \textbf{0.5961±0.0489} & \textbf{0.6186±0.0503}  \\ 
 & miRNA: multi-modal & 0.7032±0.0385 & 0.9212± 0.0096  &  0.5821± 0.0354& 0.6021± 0.0335\\ \cmidrule(lr){2-6}
 & RPPA: unimodal & 0.7546±0.0295 & \textbf{0.9433±0.0126} &0.6460±0.0257&0.6741±0.0208 \\
 & RPPA: multi-modal& \textbf{0.7628±0.0356} & 0.9432±0.0171  & \textbf{0.6576±0.0328} & \textbf{0.6843±0.0295} \\ \midrule
  \textbf{mRNA+miRNA}& mRNA: unimodal &  0.8895±0.0238& 0.9682±0.0096  & \textbf{0.8349 ±0.0366}& \textbf{0.8522±0.0359}  \\ 
 & mRNA: multi-modal & \textbf{0.8911± 0.0092}& \textbf{0.9686±0.0089}   & 0.8280 ± 0.0224& 0.8440±0.0249 \\  \cmidrule(lr){2-6}
 & miRNA: unimodal & \textbf{0.7696 ±0.0249} & 0.9429± 0.0087& \textbf{0.6755±0.0386} & \textbf{0.7040±0.0413}\\
 & miRNA: multi-modal & 0.7675±0.0425& \textbf{0.9437 ± 0.0134} & 0.6650± 0.0531& 0.6955± 0.0502\\ \midrule
  \textbf{mRNA+RPPA}& mRNA: unimodal  & 0.8802  ± 0.0219 & 0.9754± 0.0087  & 0.8199 ± 0.0297& 0.8318± 0.0317 \\ 
 & mRNA: multi-modal & \textbf{0.8969 ± 0.0165} & \textbf{0.9757± 0.0094} & \textbf{0.8378± 0.0270} & \textbf{0.8501± 0.0281} \\ \cmidrule(lr){2-6}
 & RPPA: unimodal & 0.7215± 0.0593 &0.9339 ±0.0240 & 0.6333± 0.0629& 0.6747±0.0633 \\
 & RPPA: multi-modal & \textbf{0.8512± 0.0303}& \textbf{ 0.9712± 0.0100}  & \textbf{0.7774±0.0460}  & \textbf{0.7987± 0.0479} \\ 
 \bottomrule
\end{tabular}
\end{table}

\subsection{Mask Token Replacement learns stronger representations than CLIP}

In our proposed architecture, the DuoFTT, the projected latent representations from each omics-specific FTT are averaged element-wise, and it is this vector which is used as input for the contrastive objective function. This is atypical of similar work using contrastive learning with multi-modal data, which has considered using the latent representation of each modality as input to the contrastive objective function, hence aligning the latent representations of the same sample but from different modalities \citep{hager2023best,radford2021learning}. We refer to such approaches as CLIP, named after the work which pioneered this approach, aligning the latent representations of image and text encoders, for successful image-from-text generation \citep{radford2021learning}. These approaches also use the NTXent loss during pretraining, but in this case the loss is given by
\begin{eqnarray}
\mathcal{L}_{CLIP} =  - \sum_{i=1}^{N} \log \left( \frac{  \exp \left( sim(\mathbf{u}_{i} , \mathbf{v}_{i} )/  \tau \right)   }{   \sum_{k=1,k\neq i}^{N}\exp\left( sim(\mathbf{u}_{i} , \mathbf{v}_{k}  )/ \tau \right)    } \right) - \log \left( \frac{  \exp \left( sim(\mathbf{v}_{i} , \mathbf{u}_{i} )/  \tau \right)   }{   \sum_{k=1,k\neq i}^{N} \exp\left( sim(\mathbf{v}_{i} , \mathbf{u}_{k}  )/ \tau \right)    } \right)
\end{eqnarray}
where $\mathbf{u}_i$ is the latent representation of one modality of instance $i$ and $\mathbf{v}_{i}$ is the latent representation of the other modality. Instead of performing contrastive loss on a slightly perturbed instance of the data, CLIP trains a model to align different modal representations of the same instance of data. There are two terms in the loss function, such that it is symmetric, to account for the denominator which forces the representations of different instances to be distanced in latent space. This style of pretraining can be applied to the DuoFTT architecture, by taking $\mathbf{u}_i$ and $\mathbf{v}_{i}$ to be the projected latent representations from each omics-specific FTT. During pretraining, samples of data with both modalities measured, are fed through the omics-specific FTTs of the DuoFTT. The latent representations are projected through linear layers, and these modality specific latents are used as the input to the NTXent loss as shown above. We compared whether stronger representations are learned using the typical CLIP style of pretraining multi-modal models, or if the MTR approach we proposed, learned more effective multi-omics representations. We compared a DuoFTT without pretraining, and a DuoFTT either pretrained with MTR or CLIP. We finetuned each of these models with $1\%$ of the training data. For both MTR and CLIP we pretrain for 200 epochs, and finetuned for a maximum of 200 epochs, using the validation loss as early stopping criteria with 10 epochs of patience. 

\begin{figure}[t]
  \centering
\includegraphics[width=0.45\linewidth]{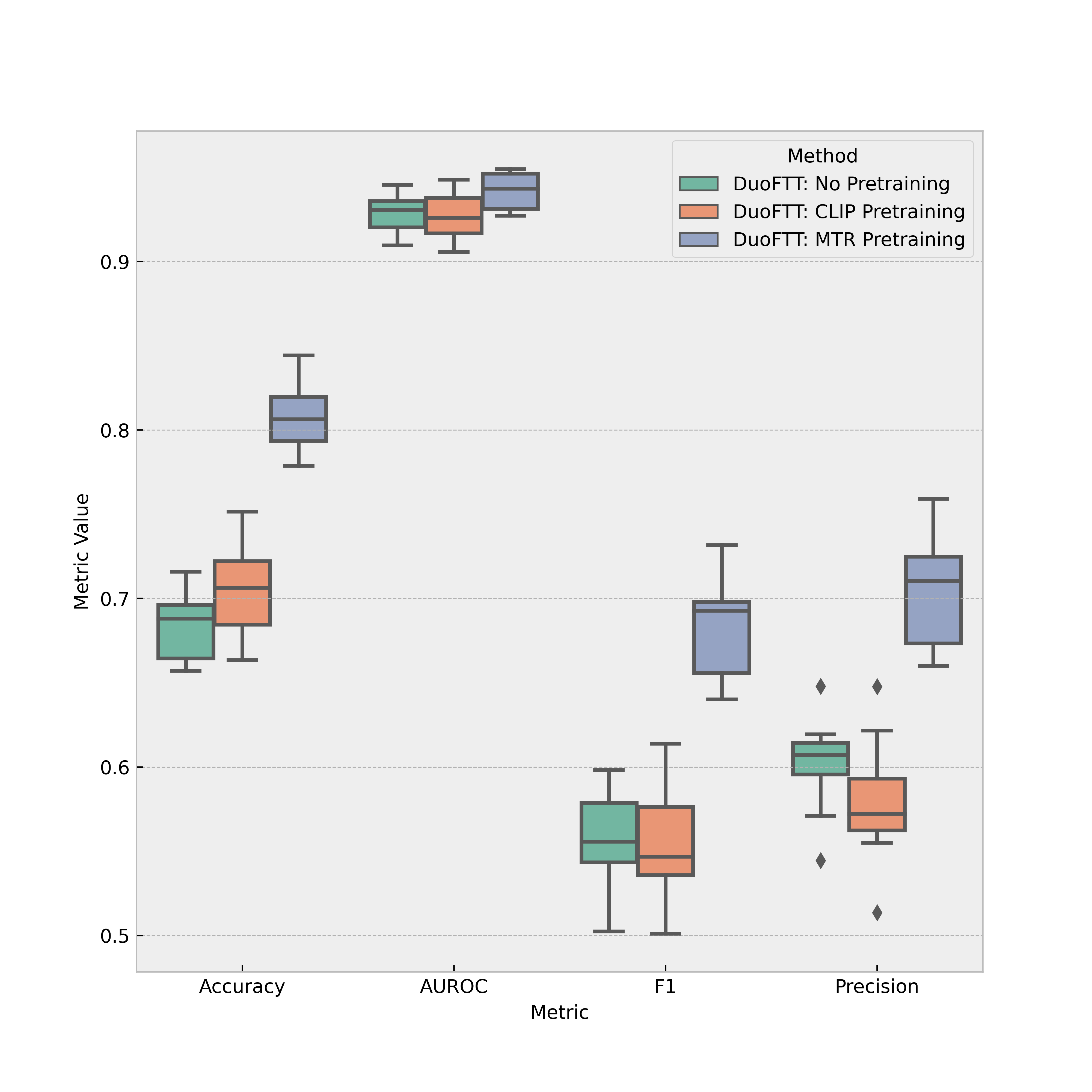} 
\includegraphics[width=0.45\linewidth]{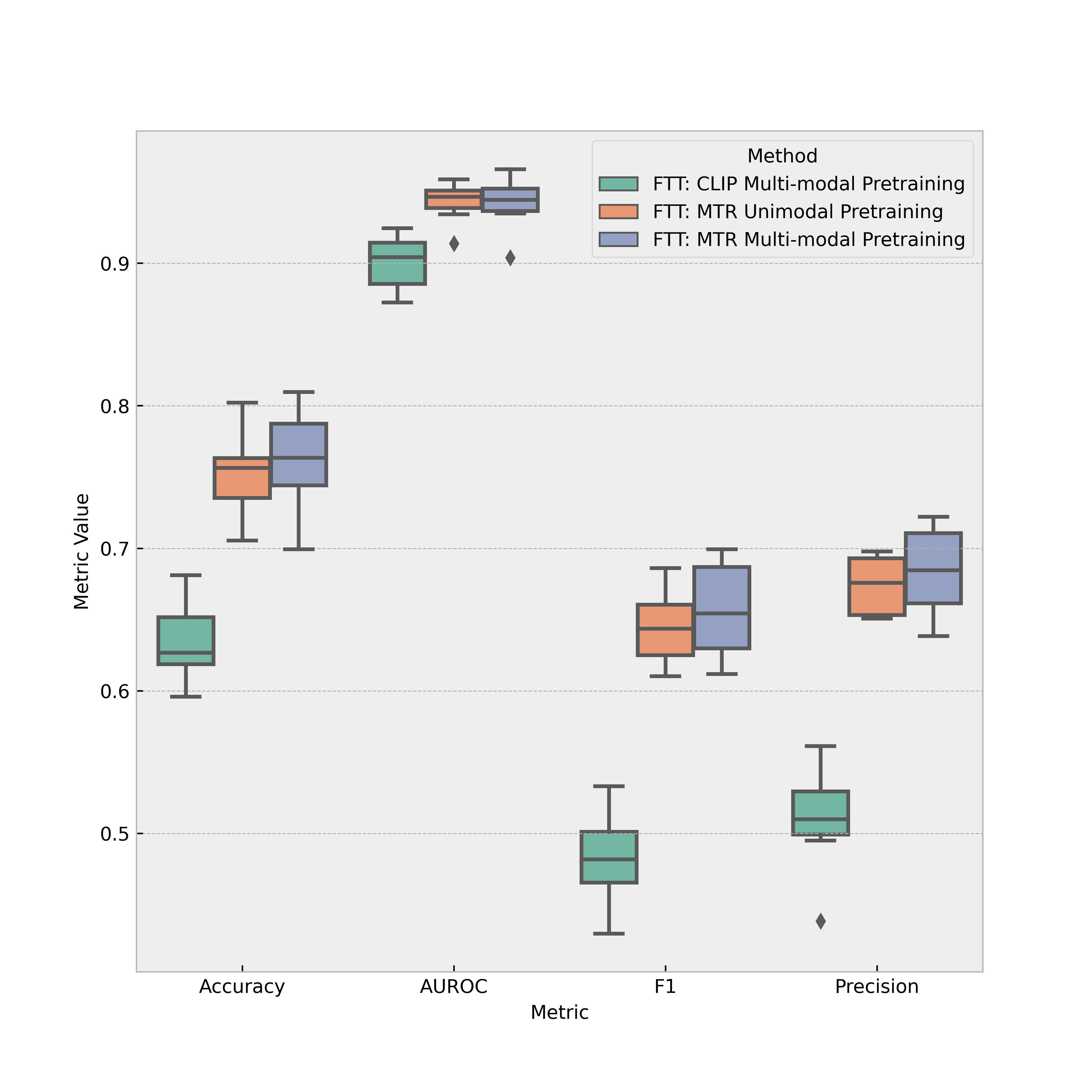}
  \caption{Results in both panels show test metrics of three different models, trained on the \textbf{miRNA+RPPA} dataset, across 10 random seeds with different data splits and weight initialisations. Left: i) DuoFTT without pretraining, ii) DuoFTT pretrained with MTR, iii)  DuoFTT pretrained with CLIP. Right: i) FTT pretrained with MTR, ii) FTT extracted from a DuoFTT pretrained with CLIP, iii) FTT extracted from a DuoFTT pretrained with MTR - each finetuned on RPPA. The test metrics shown are the overall accuracy, and macro-averaged AUROC, F1 score, and precision.  Outliers are represented by diamonds.}
\label{fig: clip vs mtr}
\end{figure}

In the left panel of Figure \ref{fig: clip vs mtr}, we find that CLIP provides an effective self-supervised signal, outperforming the DuoFTT without pretraining. However, across all test metrics, MTR is significantly more effective than CLIP. CLIP was designed, in particular, for modality-to-modality pipelines, such as generating images from text, by training multi-modal models on image-caption pairs. Hence, perhaps it is not well suited to the cancer classification task, despite earlier work suggesting this as a pretraining scheme for classification tasks \citep{hager2023best}. Hence, we hypothesised that pretraining via CLIP may provide a stronger signal for cross-omics learning. To assess this, we pretrained the DuoFTT using either MTR or CLIP and then extracted the omics-specific modules, and finetuned these on the cancer classification task, to evaluate whether multi-modal pretraining with CLIP led to stronger unimodal predictions. All models were pretrained for 200 epochs, and finetuned for a maximum of 200 epochs with validation loss as early stopping criteria with 10 epochs of patience. In the right panel of Figure \ref{fig: clip vs mtr} we show that CLIP does not provide an effective cross-omics signal for RPPA data, performing worse than a model pretrained and finetuned on RPPA only. In Appendix \ref{app: further results} we show that across each of the datasets we find that MTR outperforms CLIP (Table \ref{tab: clip vs mtr joint summary}), and that models pretrained via CLIP did not show evidence of cross-omics learning (Table \ref{tab: clip vs mtr cross summary}). 

\subsection{Matched data are not necessary for self-supervised pretraining}

Typically, multi-modal models are trained using large purposely curated multi-modal datasets. These datasets are comprised of samples which contain data from each modality, referred to as matched data. Matched samples can be passed through multi-modal architectures in a feed-forward manner in training. However, in some domains, collecting matched samples can be prohibitively difficult. In such cases, it is desirable to train a model with unmatched samples, where we collect a large amount of relevant data from each modality, but that are not necessarily from the same samples. This data can not be passed through a multi-modal model in a feed-forward manner, unless one imputes all features from the missing modalities, which is likely to lead to poor generalisation. Recent work has shown that it is possible to pretrain multi-modal models with unmatched data, and use a small number of matched samples for multi-modal prediction. The modularity of the DuoFTT means that we can pretrain an FTT using samples with features from a single omics only, and then use this pretrained FTT as the omics-specific FTT in the DuoFTT architecture. Hence, we can pretrain each omics-specific module in the DuoFTT, individually, using samples which have data from those modalities only.

To assess to what extent the DuoFTT could be pretrained with unmatched samples, we compared three different models. As baselines, we considered a DuoFTT without pretraining, and a DuoFTT that was pretrained with MTR using matched samples, as in the previous sections. We compared these baselines to a DuoFTT where each omics-specific FTT was pretrained individually, using unimodal samples from each omics. We refer to this as unmatched pretraining. In this case, each FTT was pretrained using MTR. To keep the comparison between these models fair, we split our dataset in a different manner to other experiments in this work. Firstly, we held out 20\% of the dataset for the test set. We then split the remaining data (90:10) into a train and validation set. We then split the training set further (99:1) into a labelled set for finetuning, and an unlabelled set. This unlabelled set was then split 50:50, into \textbf{Set 1} and \textbf{Set 2}.

During unmatched pretraining, when using a dataset comprised of features from \textbf{Omics 1} and  \textbf{Omics 2}, we pretrained the FTT specific to \textbf{Omics 1} with samples from \textbf{Set 1} and the FTT specific to \textbf{Omics 2} with samples from \textbf{Set 2}. During pretraining with matched samples, the DuoFTT is pretrained using all features with samples from \textbf{Set 1}. All DuoFTTs we considered in this experiment were finetuned using the same labelled splits. Both the DuoFTTs pretrained with matched and unmatched samples, were pretrained with samples from \textbf{Set 1}. However, the DuoFTT pretrained with matched samples received additional information, by seeing features from \textbf{Omics 2} from samples in \textbf{Set 1} during training. This is in contrast to the DuoFTT pretrained with unmatched samples, which was trained with features from \textbf{Omics 2} with samples in \textbf{Set 2}. By splitting the data in this way, the models can be fairly compared. Whether it was pretrained with matched or unmatched samples, both FTTs were trained using the \textit{same number} of samples, but differed in where multi-omics information was derived. For matched and unmatched pretraining, we trained for 200 epochs. All models were finetuned for 200 epochs using validation loss as an early stopping criteria with 10 epochs patience. 

\begin{figure}[t]
  \centering
\includegraphics[width=0.55\linewidth]{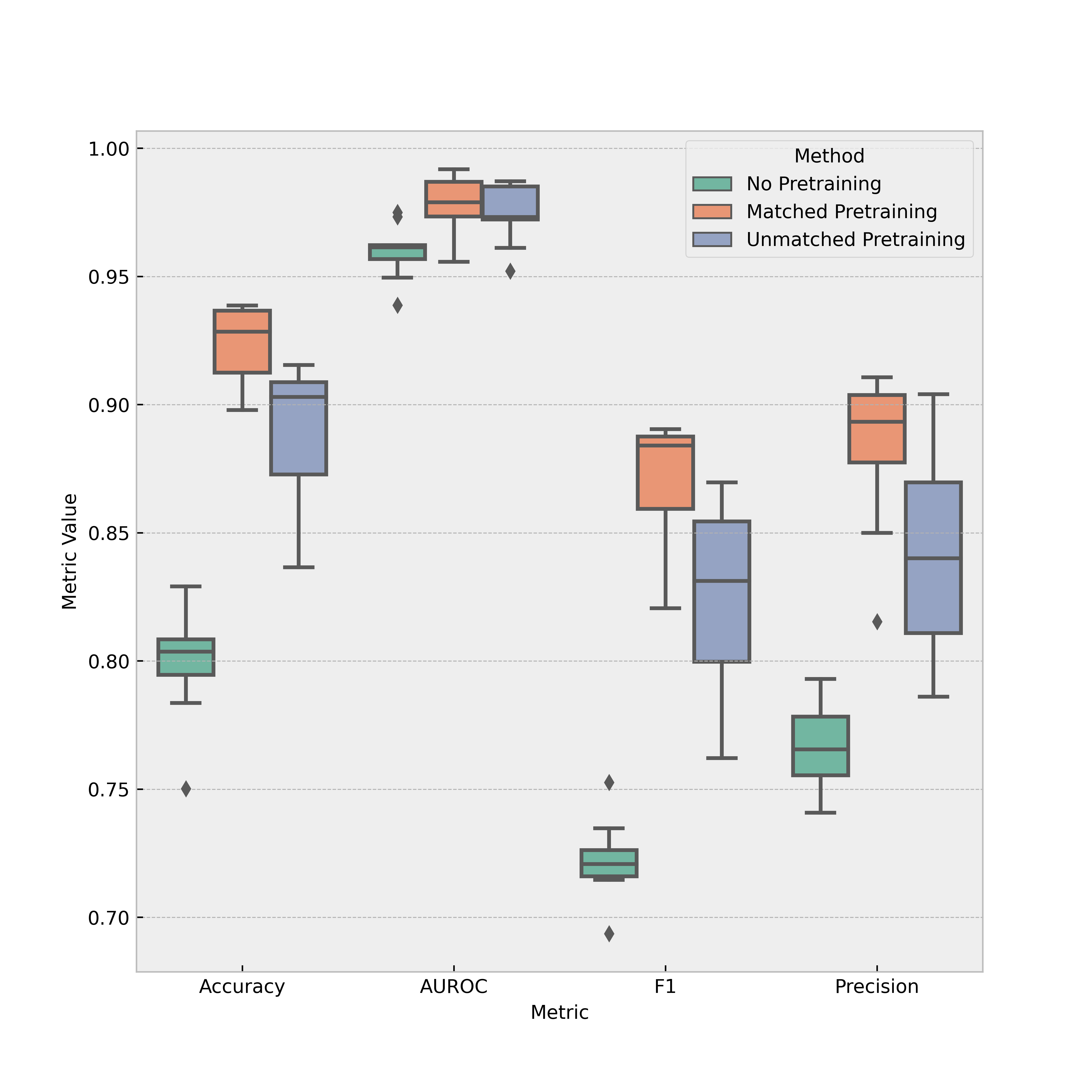}
  \caption{Test metrics of 3 models: i) DuoFTT finetuned on mRNA and RPPA, ii) DuoFTT pretrained and finetuned with matched samples of mRNA and RPPA, iii)  DuoFTT pretrained with unmatched samples of mRNA and RPPA, finetuned with matched samples of mRNA and RPPA. The test metrics shown are the overall accuracy, and macro-averaged AUROC, F1 score, and precision.  Results are shown for 10 random seeds with different data splits and weight initialisations. Outliers are represented by diamonds.}
\label{fig: unmatched vs matched}
\end{figure}

Figure \ref{fig: unmatched vs matched} shows that both pretraining schemes provide an effective self-supervised signal, and that they out perform an identical model without pretraining. However, there is a significant difference in the performance of the DuoFTTs that are pretrained with either matched or unmatched samples. Pretraining with matched samples is shown to have significantly higher performance than unmatched pretraining. This suggests that multi-modal models benefit from multi-modal samples. However, this should not rule out pretraining with unmatched samples, as this figure suggests that with a low amount of matched samples for finetuning, but large amount of unmatched unlabelled data from each modality, MTR provides an effective method to improve predictive performance with very few labels. We show in Appendix \ref{app: further results}, Table \ref{tab: matched unmatched summary}, that these results are consistent across each of the datasets we considered.

\subsection{Joint training alleviates need for multi-omics self-attention matrix}

The ability to train with unmatched samples, and to improve unimodal predictions via multi-modal pretraining relies on a \textit{modular} architecture, the network is comprised of several modules that handle each modality individually, while being available to train on a shared loss function. However, if features of multiple modalities can be properly tokenised, there is is nothing to prevent one creating a multi-modal transformer which passes all tokenised features from all modalities through a large self-attention mechanism. Indeed, one may expect that the expressivity of this model is greater, as this model would attend to cross-modal features, and explicitly model correlations between input features across both modalities. To assess whether there is a loss of expressivity by opting for a separable architecture, we compared the performance of a single FTT that takes features from both omics as input, with a DuoFTT. We finetuned both models on $1\%$ of training data for a maximum of 200 epochs, using validation loss as early stopping criteria with 10 epochs of patience.

In Table \ref{tab: duo vs single} we find that when using default parameters, the DuoFTT performs better than the multi-modal FTT across all datasets. We found that this result was consistent after HPO, as shown in Table \ref{tab: duo vs single hpo}. This suggests that in the low-label regime, an attention-based multi-modal model can benefit from a late-fusion approach, due to the smaller parameter numbers (and hence memory) required, and the better generalisation that such models may have.

\begin{table}[t]
  \caption{Comparison of test metrics of two models: i) a DuoFTT without pretraining, ii) a FTT with features from both omics used as input.   The test metrics shown are the overall accuracy, and macro-averaged AUROC, F1 score, and precision.  Test metrics are averaged over 10 seeds with (±) indicating the standard deviation. Best performance in \textbf{bold}.  }
  \label{tab: duo vs single}
   \small
\begin{tabular}{lccccccl}\toprule
& \multicolumn{5}{c}{$1\%$ training data} 
\\\cmidrule(lr){3-6}
    &       & Accuracy & AUROC & F1   & Precision  \\ \midrule
 \textbf{miRNA + RPPA}& FTT  &  0.6266±0.0238 & 0.8999±0.0120 &0.4921±0.0287& 0.5296±0.0284  \\ 
 & DuoFTT & \textbf{0.6836±0.0208 } & \textbf{0.9272±0.0132}  & \textbf{0.5554 ±0.0310} & \textbf{0.5988±0.0320} \\\midrule
  \textbf{mRNA+miRNA}& FTT  &  0.7852±0.0253 & 0.9519±0.0083 & \textbf{0.6879±0.0318} & 0.7335±0.0321  \\ 
 & DuoFTT & \textbf{0.7855±0.0229}& \textbf{0.9542±0.0082}  & 0.6800±0.0315& \textbf{0.7350±0.0360} \\ \midrule
  \textbf{mRNA+RPPA}& FTT  &  0.7761±0.0272 & 0.9536±0.0103 &0.6792±0.0408& 0.7194±0.0361  \\ 
 & DuoFTT & \textbf{0.7832±0.0247}& \textbf{0.9637±0.0103}  & \textbf{0.6937 ± 0.0423} & \textbf{0.7466± 0.0354} \\ 
 \bottomrule
\end{tabular}
\end{table}

\begin{table}[t]
  \caption{Comparison of test metrics of four models after hyperparameter optimisation:  i) a DuoFTT without pretraining, ii) a FTT with features from both omics used as input.   The test metrics shown are the overall accuracy, and macro-averaged AUROC, F1 score, and precision.  Test metrics are averaged over 10 seeds with (±) indicating the standard deviation. Best performance in \textbf{bold}.  }
  \label{tab: duo vs single hpo}
   \small
\begin{tabular}{lccccccl}\toprule
& \multicolumn{5}{c}{$1\%$ training data} 
\\\cmidrule(lr){3-6}
    &       & Accuracy & AUROC & F1   & Precision  \\ \midrule
 \textbf{miRNA + RPPA}& FTT  &  0.8072 ± 0.0224 & 0.9595 ±0.0128 & 0.7031 ± 0.0373 & 0.7338 ± 0.0377  \\ 
 & DuoFTT & \textbf{0.8296 ± 0.0179} & \textbf{0.9656 ± 0.0107} & \textbf{0.7274 ± 0.0334 } & \textbf{0.7546 ± 0.0310}  \\ \midrule
  \textbf{mRNA+miRNA}& FTT  &  0.8594 ± 0.0248 & 0.9677 ± 0.0068  & 0.7979 ± 0.0295 &0.8285 ± 0.0246  \\ 
 & DuoFTT &\textbf{ 0.8661 ± 0.0194} & \textbf{0.9698 ± 0.0086 } & \textbf{0.8005 ± 0.0220 }& \textbf{0.8313 ± 0.0198 } \\ \midrule
  \textbf{mRNA+RPPA}& FTT  & 0.8599 ± 0.0214   & 0.9750 ± 0.0076 & 0.7908 ± 0.0364 & 0.8143 ± 0.0344 \\ 
 & DuoFTT & \textbf{0.8844 ± 0167} & \textbf{0.9805 ± 0.0087}  & \textbf{0.8213 ± 0.0399} & \textbf{0.8466 ± 0.0411} \\
 \bottomrule
\end{tabular}
\end{table}

\section{Discussion}

Contrastive self-supervision, using MTR as an augmentation to generate pairs of latent representations to align, has here been shown to learn effective multi-omics representations. Models pretrained in this fashion showed superior performance in the low-label regime when compared with fully supervised methods such as GBDTs, and a self-supervised model with a reconstructive objective. Throughout this work we typically considered a small fraction of the training set (1\%) to be labelled, when demonstrating the benefit of self-supervised pretraining. Furthermore, for each of the datasets considered, the number of samples used during finetuning was $\mathcal{O}(10^{1})-\mathcal{O}(10^{2})$. Since the number of classes in these datasets was of order $\mathcal{O}(10^{1})$ this can be seen as effective few-shot learning. Acquiring labels can be prohibitively expensive in the biomedical domain, where labels may be acquired via expensive clinical trials or query by a domain expert. On the other hand, large volumes of omics data are generated by high throughput sequencing technologies. Hence, leveraging the most out of this unlabelled data is crucial in many practical settings. We have demonstrated that contrastive learning via MTR is one such approach to address this issue. 

A limitation of our work is that our model has been pretrained and finetuned on data from the same dataset. Conversely, self-supervised pretraining in computer vision and NLP has shown excellent transfer learning properties, where models are pretrained on large datasets are then finetuned on smaller, domain-specific datasets. Batch effects, where factors of a non-biological origin, such as discrepancies in the performance of measuring instruments, lead to measurable differences between batches of samples, may hinder transfer learning in the biomedical domain. However, a particular example of batch effects associated with sequencing technologies is the dropout of features measured between batches, which in this work we have shown can be mitigated with contrastive pretraining. We report little change in the performance of our pretrained model, even with a dropout of 75\% of features at test time. An interesting extension of our work would be to see to what extent contrastive learning can be used to mitigate batch effects across different public datasets by, for example, pretraining a model using data from TCGA and then finetuning using data from the Cancer Cell Line Encyclopaedia \citep{barretina2012cancer}.  

Our work demonstrates the benefits of using a modular architecture for multi-omics integration, with a separate FTT that handles each omics individually, and integrates information from different omics via the late-fusion of latent features from each FTT. This comes at the cost of not directly modelling correlations between input features from different modalities (although they will be correlated at the latent stage, and via back-propagation). However, for our bimodal model which considers two modalities each with 200 features, the computational complexity, bottlenecked by the self-attention in the transformer layers, is \textit{halved}, by considering two FTTs as oppose to one large multi-modal FTT. Besides the reduction of computational complexity, there is also a reduced risk of overfitting via over-parameterisation, as demonstrated by the superior performance of the DuoFTT compared with a FTT with features from two omics given as input. 

We have shown that there are additional benefits of a modular architecture when considering self-supervised pretraining. We reported cross-modal learning with omics data; multi-modal pretraining with mRNA and RPPA data allowed us to extract a FTT from our bimodal model that, when finetuned on RPPA data alone, performed better than the same FTT pretrained on the same RPPA data. While this phenomenon was only significant for mRNA and RPPA data, it does demonstrate that multi-modal pretraining with multi-omics data can lead to more effective single-omics models. In a practical setting, this may be useful when one has a lot of multi-omics data, but only a few unimodal samples that are labelled for some new task that one is interested in. A modular architecture also allowed us to pretrain each FTT individually, before finetuning in a joint manner. This is useful for datasets with plenty of unlabelled, unimodal data from each modality, but only a few multi-modal samples with relevant labels. While we found that multi-modal pretraining was more effective than unmatched pretraining, the self-supervised pretraining with unmatched samples did improve model performance in the low-label regime, relative to a baseline without pretraining. Notably, we have demonstrated each of these results with a modest amount of data, with at most $\mathcal{O}(10^{4})$ samples. 

In this work we limited ourselves to a multi-class classification task, using a moderately imbalanced dataset, and demonstrated that contrastive pretraining with MTR is effective in both unimodal and multi-modal settings. It remains an open question whether the latent representations learned via contrastive learning are effective for other tasks, such as regression, or even modality-to-modality pipelines i.e predicting RPPA expression from mRNA expression data. Models such as DALL-E \citep{ramesh2021zero,ramesh2022hierarchical} use contrastive learning with a CLIP loss function, aligning representations from image and text encoders, to produce a shared latent space which allows for effective translation from text to image, such that these models can be used to generate images from text. In our work, we have demonstrated that CLIP is an effective pretraining loss function, but that it is inferior to contrastive learning with MTR, and hypothesise that this is due to the diversity of data seen during training when applying random masking to input samples. Indeed it is a limitation of CLIP based approaches, that they rely on large diverse sets of matched samples for effective pretraining. Perhaps multi-modal contrastive pretraining with MTR can be used to build encoders suitable for modality-to-modality pipelines with limited matched samples. This is a multi-output regression problem, with typically high dimensional output, a far more challenging setting, the difficulty of which will only be exacerbated by limitations on the number of matched samples. 

Ultimately, we have shown for medium sized datasets, that the FTT can be robustly pretrained in both unimodal and multi-modal settings. We can achieve effective performance from very few labels per class in a multi-class classification problem, with a moderately imbalanced dataset. Crucially, contrastive self-supervised pretraining has been demonstrated to have a number of useful applications to datasets hindered by the presence of batch effects or constrained by the number of labels, or modalities, present during training.

\begin{ack}
The results shown here are in whole or part based upon data generated by the TCGA Research Network: https://www.cancer.gov/tcga.
\end{ack}

\bibliography{sslomo}

\pagebreak
\appendix

\section{Details on access to data} \label{app: data}
We accessed data from the TCGA Pan-Cancer Atlas via the UCSC Xena Platform \citep{goldman2020visualizing}. Data was downloaded from the \href{https://xenabrowser.net/datapages/?cohort=TCGA\%20Pan-Cancer\%20(PANCAN)&removeHub=http\%3A\%2F\%2F127.0.0.1\%3A7222}{Pan-Cancer Hub}\footnote{https://xenabrowser.net/datapages/} with our datasets (in bold) corresponding to the following subsets listed on the Xena platform (shown in quotation marks):
\begin{itemize}
\item \textbf{mRNA}: ``Batch effects normalized mRNA data"\\
\item \textbf{miRNA}:  ``Batch effects normalized miRNA data" \\
\item \textbf{RPPA}: ``RPPA". \\ 
\end{itemize}

\section{Further tables and figures} \label{app: further results}

\setcounter{figure}{0}
\renewcommand{\thefigure}{\arabic{section}.\arabic{figure}}

\setcounter{table}{0}
\renewcommand{\thetable}{\thesection.\arabic{table}}

\begin{table}[hp]
  \caption{Comparison of performance on the cancer classification task of 4 models i) FTT ii) FTT with contrastive pretraining via MTR iii) CatBoost and iv) XGBoost. The test metrics shown are the overall accuracy, and macro-averaged AUROC, F1 score, and precision.  Test metrics are averaged over 10 seeds with (±) indicating the standard deviation. Best performance in \textbf{bold} and second best is \underline{underlined}.  }
  \label{tab: GBDT comparison all omics}
  \small
\begin{tabular}{lccccccl}\toprule
& \multicolumn{5}{c}{$1\%$ training data} 
\\\cmidrule(lr){3-6}
    &       & Accuracy & AUROC & F1   & Precision  \\ \midrule
 \textbf{mRNA}&FTT  & \underline{0.7712 ± 0.0281} &0.9286 ± 0.0157 &\underline{0.6413 ±0.0378}& 0.6796 ± 0.0310\\ 
 &MTR & \textbf{0.8684 ± 0.0157} & \underline{0.9469± 0.0128} & \textbf{0.7729 ±0.0295} & \textbf{ 0.7892±0.0322 } \\
 &CatBoost & 0.6445± 0.0360 &\textbf{ 0.9614±0.0061 } & 0.5066 ± 0.0298 & \underline{ 0.7016±0.0481 } \\
 &XGBoost &0.5828 ± 0.0272 & 0.8851 ± 0.0190 & 0.4490±0.0226 &0.4824 ±0.0274\\ \midrule
  \textbf{miRNA}&FTT  & \underline{0.6222± 0.0215} &0.8829 ±  0.0178&\underline{0.4918±0.0326} & 0.5272 ± 0.0375 \\
 &MTR & \textbf{ 0.7640± 0.0249} & \underline{0.9222 ±0.0158 } & \textbf{ 0.6584± 0.0317} & \textbf{ 0.6770± 0.0330} \\
 &CatBoost & 0.5652±0.0181  &\textbf{0.9372±0.0070} & 0.4403±0.0242  & \underline{0.6038±0.0319} \\
 &XGBoost  & 0.4812±0.0277  & 0.8322±0.0141 & 0.3628±0.0206  & 0.4020 ±0.0307 \\ \midrule
  \textbf{RPPA}&FTT  &\underline{0.5090±0.0276} & 0.8413±0.0203 &\underline{0.3865 ±0.0232} & 0.4209±0.0325 \\
 &MTR & \textbf{ 0.7297±0.0327 } & \textbf{0.9201 ± 0.0161} & \textbf{ 0.6201± 0.0411} & \textbf{0.6546 ± 0.0406} \\
 &CatBoost & 0.4183± 0.0455 &\underline{ 0.8691 ± 0.0204} & 0.2993 ± 0.0312 & \underline{ 0.4285±0.0569 } \\
 &XGBoost & 0.3635 ±  0.0363& 0.7799± 0.0213 &0.2607  ± 0.0295 & 0.2824± 0.0302\\
 \bottomrule
\end{tabular}
\end{table}

\begin{table}[hp]
  \caption{Comparison of performance on the cancer classification task of 4 models i) FTT trained with PCA  and ii) FTT  iii) CatBoost and iv) XGBoost, each trained without PCA. Models trained without PCA are indicated with an asterisk. The test metrics shown are the overall accuracy, and macro-averaged AUROC, F1 score, and precision.  Test metrics are averaged over 10 seeds with (±) indicating the standard deviation. Best performance in \textbf{bold} and second best is \underline{underlined}.}
  \label{tab: raw GBDT comparison}
  \small
\begin{tabular}{lccccccl}\toprule
& \multicolumn{5}{c}{$1\%$ training data} 
\\\cmidrule(lr){3-6}
    &       & Accuracy & AUROC & F1   & Precision  \\ \midrule
 \textbf{mRNA}&FTT  & \textbf{0.7712 ± 0.0281} & \underline{0.9286 ± 0.0157} &\textbf{0.6413 ±0.0378}& \textbf{0.6796 ± 0.0310}\\
&FTT* &  -  & -  & - & - \\ 
 &CatBoost* & \underline{0.6743 ± 0.0267} & \textbf{0.9586 ± 0.0117} & \underline{0.5254 ± 0.0330} & \underline{0.6773 ± 0.0674} \\
  &XGBoost* & 0.5782 ± 0.0430& 0.8745 ± 0.0228  & 0.4547± 0.0350 & 0.4739 ± 0.0331 \\ \midrule
  \textbf{miRNA}&FTT  & 0.6222± 0.0215 &0.8829 ±  0.0178& 0.4918 ± 0.0326 & 0.5272 ± 0.0375 \\
  &FTT* &  \textbf{0.7242 ± 0.0188} & \underline{0.9166 ± 0.0144} & \textbf{0.6131 ± 0.0262} & \underline{0.6436 ± 0.0275} \\ 
  &CatBoost* & \underline{0.6342 ± 0.0302} & \textbf{0.9560 ± 0.0065} & \underline{0.5257± 0.0325} & \textbf{0.6500 ± 0.0537} \\
  &XGBoost* & 0.5761 ± 0.0226 & 0.9058± 0.0090& 0.4670 ± 0.0291 & 0.5155  ± 0.0310 \\ \midrule
  \textbf{RPPA}&FTT  & 0.5090 ± 0.0276 & 0.8413±0.0203 & 0.3865 ± 0.0232 & 0.4209 ± 0.0325 \\
  &FTT* & \textbf{0.6956 ± 0.0169} & \underline{0.9133 ± 0.0145} & \textbf{0.5623 ± 0.0180} & \textbf{0.5907 ± 0.0245 }\\ 
  &CatBoost* & \underline{0.5345 ± 0.0488}  & \textbf{0.9240 ± 0.0146} & \underline{0.4144 ± 0.0448} & \underline{0.5213 ± 0.0598} \\
  &XGBoost* & 0.4808 ± 0.0379  & 0.8664 ± 0.0228 & 0.3773± 0.0348 & 0.4039 ± 0.0377\\
 \bottomrule
\end{tabular}
\end{table}

\begin{table}[hp]
  \caption{For each dataset, we compare the DuoFTT with i) no pretraining, ii) pretraining using CLIP loss applied to latent of each FTT and iii) pretraining using MTR applied to the average latent of each FTT.  The test metrics shown are the overall accuracy, and macro-averaged AUROC, F1 score, and precision. Test metrics are averaged over 10 seeds with (±) indicating the standard deviation. Best performance in \textbf{bold} and second best is \underline{underlined}.  }
  \label{tab: clip vs mtr joint summary}
  \small
\begin{tabular}{lccccccl}\toprule
& \multicolumn{5}{c}{$1\%$ training data} 
\\\cmidrule(lr){3-6}
    &       & Accuracy & AUROC & F1   & Precision  \\ \midrule
 \textbf{miRNA + RPPA}& No Pretraining  & 0.6838 ± 0.0204& \underline{0.9286± 0.0122}& \underline{0.5566±0.0294}& \underline{0.6013± 0.0280} \\ 
 & CLIP & \underline{0.7069±0.0313}& 0.9248±0.0125  & 0.5526 ± 0.0358& 0.5773±0.0388 \\
 & MTR & \textbf{0.8067±0.0202} & \textbf{0.9418± 0.0111} & \textbf{0.6830±0.0295}&\textbf{0.7032±0.0325} \\ \midrule
  \textbf{mRNA+miRNA}& No Pretraining  & 0.7823 ± 0.0264& 0.9546± 0.0075& 0.6759±0.0365& 0.7312± 0.0394 \\ 
 & CLIP & \underline{0.8281±0.0287}& \underline{0.9582±0.0050}  & \underline{0.7307 ± 0.0474} &\underline{0.7589 ±0.0504} \\
 & MTR & \textbf{0.8859± 0.0139}& \textbf{0.9702±0.0100} & \textbf{0.8180±0.0284}&\textbf{0.8377±0.0269} \\ \midrule
  \textbf{mRNA+RPPA}& No Pretraining  &  0.7994±0.0211& 0.9600± 0.0105& 0.7219±0.0152& \underline{0.7666±0.0161}  \\ 
 & CLIP & \underline{0.8320±0.0386} &\underline{ 0.9674 ± 0.0095} &\underline{0.7356±0.0546}  & 0.7584 ± 0.0501  \\
 & MTR & \textbf{0.9282±0.0132}  &\textbf{0.9848±0.0086} & \textbf{0.8784±0.0268}&\textbf{0.8905±0.0300} \\ 
 \bottomrule
\end{tabular}
\end{table}

\begin{table}[hp]
  \caption{Comparison of the performance of an FTT, extracted from a DuoFTT pretrained with either CLIP or MTR, on the cancer classification task. For each omics in each dataset, we compare the test metrics of three models finetuned on data from that omics: i) a FTT pretrained on the same omics ii) a FTT extracted from a DuoFTT pretrained via MTR (M Cross) and iii) a FTT extracted from a DuoFTT pretrained via CLIP (C Cross). The test metrics shown are the overall accuracy, and macro-averaged AUROC, F1 score, and precision.  Test metrics are averaged over 10 seeds with (±) indicating the standard deviation. Best performance in \textbf{bold} and second best is \underline{underlined}.  }
  \label{tab: clip vs mtr cross summary}
 \small
\begin{tabular}{lccccccl}\toprule
& \multicolumn{5}{c}{$1\%$ training data} 
\\\cmidrule(lr){3-6}
    &       & Accuracy & AUROC & F1   & Precision  \\ \midrule
 \textbf{miRNA + RPPA}&  miRNA: MTR &  \textbf{0.7159±0.0390} & \textbf{0.9232±0.0137} & \textbf{0.5961±0.0489} & \textbf{0.6186± 0.0503}\\ 
 & miRNA: M Cross & \underline{0.6858 ± 0.0254}& \underline{0.9011±0.0138 }& \underline{0.5674±0.0336}& \underline{0.5937±0.0327} \\
 & miRNA: C Cross & 0.6410±0.0333& 0.8981±0.0153 & 0.4920±0.0331&0.5145±0.0370 \\ \cmidrule(lr){2-6}
&  RPPA: MTR &  \underline{0.7546± 0.0295}& \textbf{0.9433±0.0126} & \underline{0.6460 ±0.0257} & \underline{0.6741± 0.0208}\\ 
 & RPPA:  M Cross & \textbf{0.7614±0.0346}& \underline{0.9408±0.0171} & \textbf{0.6536±0.0336}& \textbf{0.6804±0.0309} \\
 & RPPA: C Cross & 0.6323±0.0268& 0.9005±0.0179 & 0.4843±0.0315&0.5129±0.0351 \\ \midrule
  \textbf{mRNA+miRNA}&  mRNA: MTR &  \underline{0.8895±0.0238} & \underline{0.9682±0.0096} & \textbf{0.8349 ±0.0366}&\textbf{0.8522 ±0.0359} \\ 
 & mRNA: M Cross & \textbf{0.8902±0.0087}& \textbf{0.9687±0.0088} & \underline{0.8273±0.0225}& \underline{0.8429±0.0251} \\
 & mRNA: C Cross & 0.8282±0.0267& 0.9571±0.0064 &0.7325 ± 0.0414& 0.7619±0.0453 \\ \cmidrule(lr){2-6}
&  miRNA: MTR & \underline{ 0.7696 ± 0.0249}& 0.9429± 0.0087& \underline{0.6755±0.0386}& \underline{0.7040±0.0413} \\ 
 & miRNA: M Cross & 0.7675±0.0425& \underline{0.9437±0.0134} & 0.6650±0.0531& 0.6955±0.0502 \\
 & miRNA: C Cross & \textbf{0.7838±0.0213}& \textbf{0.9490±0.0100} & \textbf{0.6787±0.0351}&\textbf{0.7047±0.0373} \\ \midrule
  \textbf{mRNA+RPPA}&  mRNA: MTR &  \underline{0.8802±0.0219} & \underline{0.9754±0.0087} & \underline{0.8199 ±0.0297}& \underline{0.8318± 0.0317} \\ 
 & mRNA: M Cross & \textbf{0.8966±0.0163}& \textbf{0.9778±0.0095} & \textbf{0.8416±0.0274}& \textbf{0.8553±0.0284} \\
 & mRNA: C Cross & 0.8086±0.0403& 0.9606±0.0106 &0.7130 ±0.0563&0.7449±0.0530 \\ \cmidrule(lr){2-6}
&  RPPA: MTR & 0.7215 ± 0.0593 & 0.9339±0.0240 & 0.6333±0.0629&\underline{0.6747 ±0.0633} \\ 
 & RPPA: M Cross & \textbf{0.8512±0.0303}& \textbf{0.9712±0.0100} & \textbf{0.7774 ±0.0460}& \textbf{0.7987±0.0479} \\
 & RPPA: C Cross & \underline{0.7417±0.0479}& \underline{0.9464±0.0116} & \underline{0.6335 ±0.0637}&0.6566±0.0634 \\ \bottomrule
\end{tabular}
\end{table}

\begin{table}[p]
  \caption{Comparison of performance on the cancer classification task of 3 models: i) a DuoFTT without pretraining, ii) a DuoFTT where each FTT has been pretrained individually, which is then finetuned with multi-modal samples (Unmatched) and iii) a DuoFTT which has been pretrained and finetuned with multi-modal samples (Matched).  The test metrics shown are the overall accuracy, and macro-averaged AUROC, F1 score, and precision. Test metrics are averaged over 10 seeds with (±) indicating the standard deviation. Best performance in \textbf{bold} and second best is \underline{underlined}.  }
  \label{tab: matched unmatched summary}
   \small
\begin{tabular}{lccccccl}\toprule
& \multicolumn{5}{c}{$1\%$ training data} 
\\\cmidrule(lr){3-6}
    &       & Accuracy & AUROC & F1   & Precision  \\ \midrule
 \textbf{miRNA + RPPA}& No Pretraining & 0.6838 ± 0.0204 &0.9286 ± 0.0122 &0.5566 ±0.0294& 0.6013± 0.0280\\ 
 & Matched & \textbf{0.8177 ± 0.0316} & \textbf{0.9605±  0.0114} & \underline{0.7062 ±0.0360} & \underline{ 0.7225±0.0303 } \\
 & Unmatched  & \underline{0.8113± 0.0208} & \underline{0.9562±0.0145} & \textbf{0.7071±0.0295}& \textbf{0.7314 ±  0.0313}\\\midrule
  \textbf{mRNA+miRNA}& No Pretraining &0.7823  ±0.0264 &0.9546 ±0.0075  & 0.6759±0.0365& 0.7312± 0.0394 \\ 
 & Matched & \textbf{0.8859± 0.0139}& \textbf{0.9702±0.0100}   & \underline{0.8180 ± 0.284}& \underline{0.8377± 0.269} \\
 & Unmatched & \underline{0.8812±0.0158} & \underline{0.9696±0.0091} &\textbf{0.8187±0.0309}& \textbf{0.8419 ±0.0326} \\ \midrule
  \textbf{mRNA+RPPA}& No Pretraining  & 0.7994 ± 0.0211& 0.9600± 0.0105 & 0.7219± 0.0152& 0.7666± 0.0161 \\ 
 & Matched & \textbf{0.9237± 0.0158} & \textbf{0.9787± 0.0106} & \textbf{0.8706 ± 0.0238}  & \textbf{0.8831± 0.0300} \\
 & Unmatched & \underline{0.8910 ± 0.0275} & \underline{0.9748± 0.0116} & \underline{0.8270 ± 0.0363} & \underline{0.8431±0.0384} \\
 \bottomrule
\end{tabular}
\end{table}

\clearpage
\section{Hyperparameter optimisation} \label{app: hpo}

\setcounter{table}{0}
\renewcommand{\thetable}{\thesection.\arabic{table}}

In Section \ref{sec: hpo} we compare the performance of several models after hyperparameter optimisation. To perform optimisation we use the software library Optuna \citep{akiba2019optuna}. We compare the performance of these models over 5 seeds. For each seed, we split our dataset, perform HPO, and then train our optimised model on our final dataset and computed test metrics. For each round of HPO, we run 100 trials of each model, each time selecting parameters from the distributions specified in Table \ref{tab: hpo optimisation}.

\begin{table}[ht]
  \caption{Parameter distributions used in hyperparameter optimisation for several models.}
  \label{tab: hpo optimisation}
\begin{tabular}{lccccl}\toprule
    Model &    Parameter   & Range &  \\ \midrule
 \textbf{FTT}& \# Layers  & $ [1,4]$  &      \\ 
 & Token dimension  & $[64,512]$  &      \\ 
 & Residual dropout & $[0,0.2]$  &      \\ 
 & Attention dropout & $[0,0.5]$  &     \\ 
 & Feed-forward network dropout & $[0,0.5]$  &      \\ 
 & Feed-forward network factor & $[\frac{2}{3},\frac{8}{3}]$  &      \\
  & Learning rate & $[1\mathrm{e}^{-5},1\mathrm{e}^{-3}]$  &      \\
   & Weight decay & $[1\mathrm{e}^{-6},1\mathrm{e}^{-3}]$   &      \\
 \midrule
  \textbf{MLP}& \# layers  & $[3,6]$  &    \\
& Layer size factor  & $[0.5,1.0]$  &    \\
& \# training epochs  & $[15,200]$  &    \\
& Batch size  & $[32,128]$  &   \\
 & Learning rate & $[1\mathrm{e}^{-4},0.5]$  &      \\
  \midrule
   \textbf{VIME}& \# MLP layers & $[1,5]$  &    \\
    & \# Nodes per layer & $[100,500]$   &      \\
      & Alpha & $[0.1,10]$   &     \\
& Mask probability  & $[0.1,0.75]$  &    \\
& Batch size  & $[50,128]$  &    \\
& \# Training epochs  & $[5,50]$  &   \\
 \midrule
  \textbf{CatBoost}& Max depth & $[3,10]$  &    \\
& Learning rate  & $[1\mathrm{e}^{-5},1.0]$  &    \\
& Bagging temperature  & $[0,1.0]$  &    \\
& L2 Leaf Regularisation  & $[1,10]$  &    \\
 & Leaf estimation iterations & $[1,10]$  &      \\
 \midrule
  \textbf{XGBoost}& Max depth & $[3,10]$  &    \\
    & Minimum child weight & $[1\mathrm{e}^{-8},1\mathrm{e}^{5}]$   &      \\
      & Subsample & $[0.5,1.0]$   &   & &   \\
& Learning rate  & $[1\mathrm{e}^{-5},1.0]$  &    \\
& Column sample by level  & $[0.5,1.0]$  &    \\
& Column sample by tree  & $[0.5,1.0]$  &    \\
& Gamma & $[1\mathrm{e}^{-8},1\mathrm{e}^{2}]$   &      \\
& Lambda & $[1\mathrm{e}^{-8},1\mathrm{e}^{2}]$   &      \\
 & Alpha & $[1\mathrm{e}^{-8},1\mathrm{e}^{2}]$   &      \\
 \midrule
 \bottomrule
\end{tabular}
\end{table}

\end{document}